\documentclass[final,1p,times]{elsarticle}
\usepackage{array}
\usepackage{amsmath,amssymb,amsfonts,amsthm}
\newtheorem{theorem}{Theorem}
\newtheorem{lemma}{Lemma}

\newtheorem{definition}{Definition}
\newtheorem{Proposition}{Proposition}
\usepackage{graphicx}
\usepackage{epstopdf}% only for final version
\usepackage{subfigure}
\usepackage{algorithm}
\usepackage{algorithmic}
\usepackage{url}
\usepackage{bm}
\usepackage{caption}
\usepackage{multirow}
\usepackage[T1]{fontenc}
\usepackage{stfloats}
\usepackage{tabularx}
\usepackage{booktabs}
\usepackage{lipsum}
\usepackage{mathrsfs}
\usepackage{amsmath}
\usepackage[numbers]{natbib}
\usepackage{ragged2e}
\usepackage{hyperref}
\usepackage{wrapfig}
\journal{Artificial Intelligence}

\begin{document}

\begin{frontmatter}

\title{Theoretically Guaranteed Distribution Adaptable Learning}
%\tnotetext[mytitlenote]{Corresponding author at: College of Science, National University of Defense Technology, Changsha, 410073, China}
%% Group authors per affiliation:
%\author{Elsevier\fnref{myfootnote}}
%\address{Radarweg 29, Amsterdam}
%\fntext[myfootnote]{Since 188.}

%% or include affiliations in footnotes:
\author[mymainaddress]{Chao Xu}
\author[mymainaddress]{Xijia Tang}
\author[mysecondaryaddress]{Guoqing Liu}
\author[mysecondaryaddress]{{Yuhua Qian}\corref{mycorrespondingauthor}}
\author[mymainaddress]{Chenping Hou\corref{mycorrespondingauthor}}
\ead{jinchengqyh@sxu.edu.cn; hcpnudt@hotmail.com}
\cortext[mycorrespondingauthor]{Corresponding authors.}
\address[mymainaddress]{College of Science, National University of Defense Technology, Changsha, 410073, China.}
\address[mysecondaryaddress]{The Institute of Big Data Science and Industry, Shanxi University, Taiyuan, 030006, Shanxi, China.}

\begin{abstract}
In many open environment applications, data are collected in the form of a stream, which exhibits an evolving distribution over time. How to design algorithms to track these evolving data distributions with provable guarantees, particularly in terms of the generalization ability, remains a formidable challenge. To handle this crucial but rarely studied problem and take a further step toward robust artificial intelligence, we propose a novel framework called Distribution Adaptable Learning (DAL). It enables the model to effectively track the evolving data distributions. By Encoding Feature Marginal Distribution Information (EFMDI), we broke the limitations of optimal transport to characterize the environmental changes and enable model reuse across diverse data distributions. It can enhance the reusable and evolvable properties of DAL in accommodating evolving distributions. Furthermore, to obtain the model interpretability, we not only analyze the generalization error bound of the local step in the evolution process, but also investigate the generalization error bound associated with the entire classifier trajectory of the evolution based on the Fisher-Rao distance. For demonstration, we also present two special cases within the framework, together with their optimizations and convergence analyses. Experimental results over both synthetic and real-world data distribution evolving tasks validate the effectiveness and practical utility of the proposed framework.
\end{abstract}

\end{frontmatter}

%\linenumbers
\section{Introduction}
\begin{figure*}[!t]
	\setlength{\abovecaptionskip}{-.1cm}
	\setlength{\belowcaptionskip}{-.1cm}
	\begin{center}
		\includegraphics[width=13cm]{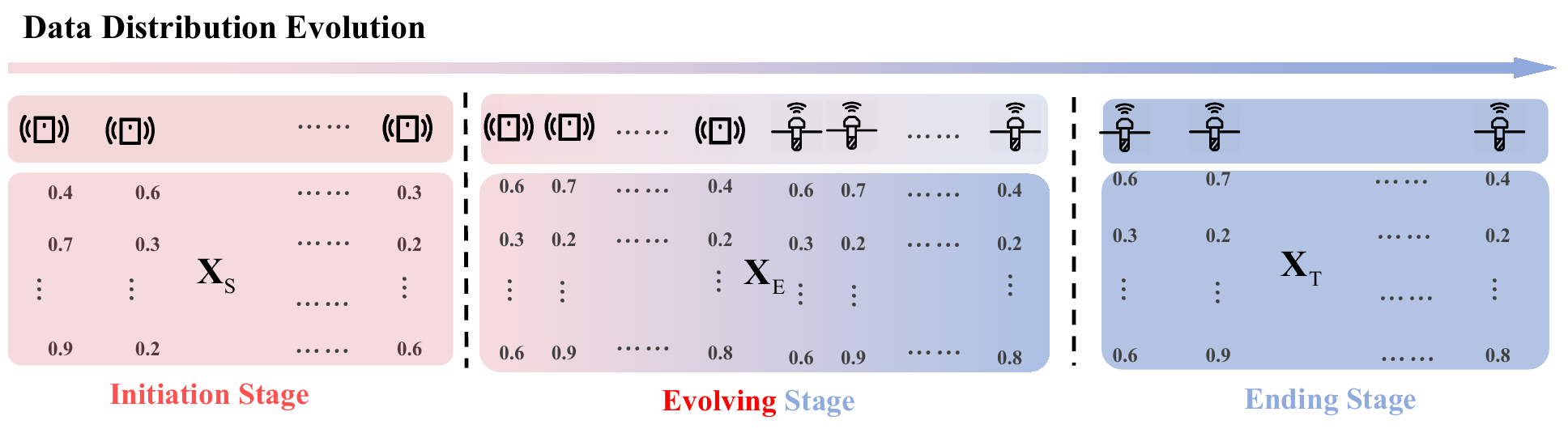}
	\end{center}
	\caption{Illustration for Data Distribution Evolving Streaming Learning. In environment monitoring task, the data distribution might be evolving gradually in the streaming data.}
	\label{figure1}
	\vskip -.2in
\end{figure*}

Machine learning, a fundamental research area of artificial intelligence, has achieved great success in many tasks, especially in supervised learning \cite{DBLP:journals/nature/LeCunBH15}. In conventional supervised learning, one of the most key assumptions is the `independent and identical distribution' assmpution, i.e., i.i.d. assumption. It requires that the training and testing data are independently and identically distributed. However, this assumption has been easily violated in many real applications \cite{DBLP:journals/ai/DeJong94,DBLP:journals/ml/Ben-DavidBCKPV10,DBLP:journals/ai/NguyenSLL17}. For example, in many open environment scenarios, data are usually accumulated over time and collected dynamically \cite{DBLP:journals/corr/abs-2206-00423,DBLP:conf/icml/Zhang0JZ20}. In particular, the distribution of these streaming data can evolve, where the source distribution gradually evolves into the target distribution. For instance, we have a self-driving agent equipped with a scene recognition system that has been trained on scenes from a stationary condition (the source distribution). When it is deployed in the real world, the environment can vary in a continually evolving way, such as from day to night or from sunshine to rain. Therefore, we expect that the agent can gradually adapt to the environment shift and perform consistently well on scenes from all environments (the evolving distributions). A similar situation also occurs in news text classification tasks, where the evolving data distribution accompanies the event development process, and the enduring popularity of new hotspots is accompanied by the gradual forgetting of old ones. In these scenarios, as shown in Fig. \ref{figure1}, the data samples come like a stream over time. Generally, the distribution evolving process could be divided into three stages, i.e., initiation, evolving, and ending stages. The evolving stage will persist for a certain duration.

Different from traditional cases, there are several challenges, e.g., the difficulties in data collection, storage, and labeling, in a dynamic environment. On one hand, due to the limitation of computational and storage resources, particularly when the data comes like a stream, it is infeasible to keep the whole data for optimization. On the other hand, due to the expensive labeling cost, there are usually only a few collected labeled data, especially within a short period. Evolving data distributions and limited labeled data pose an obstruction to traditional classification methods. Specifically, evolving data distribution violates the i.i.d. assumption and the limited labeled data are not enough to support traditional methods to train a satisfactory model. The presence of these obstacles renders traditional approaches for evolving data distribution classification tasks.

In the literature, transfer learning can tackle the problem of distribution change. Mainstream domain adaptation methods are tailored to adapting discrete source and target domains \cite{DBLP:journals/ml/Ben-DavidBCKPV10,DBLP:conf/nips/Ben-DavidBCP06,DBLP:journals/corr/abs-2010-03978,DBLP:conf/ijcai/0001LLOQ21}. This line of works rely on a domain discriminator to distinguish two discrete domains, and cannot be applied to evolving target domain directly. Despite we can discretize the evolving data distribution, the limited availability of labeled data at each moment continues to impede existing domain adaptation methods because these methods require fully labeled source domain data to transfer knowledge to the unlabeled target domain. 

Besides, the violation of i.i.d. assumption also poses an obstruction to model reuse, which is another way for dynamic data. Mainstream model reuse methods are based on the i.i.d. assumption \cite{DBLP:conf/icml/YeZ0Z18, aaai/YangZFJZ17}. With model reuse, the new-coming limited labeled data can make the model sharper. The evolving data distribution limits the direct model reuse between feature domains. There are also some heterogeneous model reuse methods. For example, the REctiFy via heterOgeneous pRedictor Mapping (REFORM) framework, proposed by Ye et al., presents a solution for effectively utilizing models across diverse features or labels \cite{pami/YeZJZ21}. Although these approaches have achieved prominent performance in their applications, they mainly deal with scenarios involving heterogeneous feature spaces, which are not consistent with our scenario. Thus, they are inapplicable to the problem at hand.

To solve these problems, we borrow the main idea from \emph{Learnware} \cite{DBLP:journals/fcsc/Zhou16a}. It presents a novel perspective towards robust modeling, which is a well-performed pre-trained learner with specifications. Two essential properties of learnware, i.e., reusable and evolvable, are emphasized in this work. We make a preliminary step towards robust modeling guided by \emph{Learnware} by proposing a novel framework named Distribution Adaptable Learning (DAL) to track the evolving data distributions. By Encoding Feature Marginal Distribution Information (EFMDI), we expand the scope of optimal transport to bridge the environmental changes and enable model reuse across diverse data distributions. It can enhance the reusable and evolvable properties of DAL in accommodating evolving data distributions. Specifically, EFMDI enables us to establish a transformation between feature sets and models, allowing the rectification of the former model through this transformation in current tasks. Moreover, since annotation is expensive, only limited annotated data samples are available, especially within a short period. Inspired by semi-supervised learning, we employ manifold regularization to extract the structural information of data to mitigate the challenges posed by limited labeled data.  It can enhance both the model performance and generalization ability.

The main contributions are four folds.

\begin{itemize}
	\item This paper makes a preliminary step towards robust modeling guided by \emph{learnware}. It contains two parts to implement the reusable and evolvable properties accordingly. We develop a new model reuse framework DAL on evolving data distribution in a dynamic environment, and propose a novel evolvability solution via bridging evolving domains, which can be reused for tasks with evolving data distribution. 
	\item The rigorous generalization ability analysis provides a solid theoretical guarantee for DAL framework. We not only analyze the generalization error bound of the local step in the evolution process, but also investigate the generalization ability associated with the entire classifier evolution trajectory.
	\item We propose a novel strategy named Encoding Feature Marginal Distribution Information (EFMDI), which expands the scope of optimal transport to bridge the environmental changes and enables model reuse across diverse data distributions. It illustrates the effectiveness of DAL in accommodating evolving data distribution.
	\item Comprehensive experimental studies demonstrate the effectiveness of our proposal on several data sets in a broad range. The experimental results indicate that our method outperforms other compared approaches in most cases. Besides, we have also utilized our method in digit recognition with a large application potentiality.
\end{itemize}

The rest of this manuscript starts with some notations in Section 2. Then, the DAL framework is presented in detail, including the overall framework and specific components. In Section 3, we present a meticulous theoretical analysis of the proposed DAL framework. In Section 4, we detail the analyses of two specific algorithms within the proposed DAL framework, together with their optimizations and convergence analyses. In Section 5, we review the closely related work about domain adaptation and model reuse. Experimental results on benchmark and real data sets are displayed in Section 6. Finally, we conclude this paper in Section 7.

\section{The DAL Problem} \label{s2}

We will give some notations at first. After that, we propose the DAL framework, together with the specific components.

\subsection{Notations and Problem Setting} \label{s2.1}

In streaming data learning, at each time, a batch of data is received where only their features are available. We are required to predict their labels. In our scenario, the data distribution of the consecutive data batches might be evolving gradually due to environmental change.

\begin{figure}[t]
	\setlength{\abovecaptionskip}{-.1cm}
	\setlength{\belowcaptionskip}{-.1cm}
	\begin{center}
		\includegraphics[width=12cm]{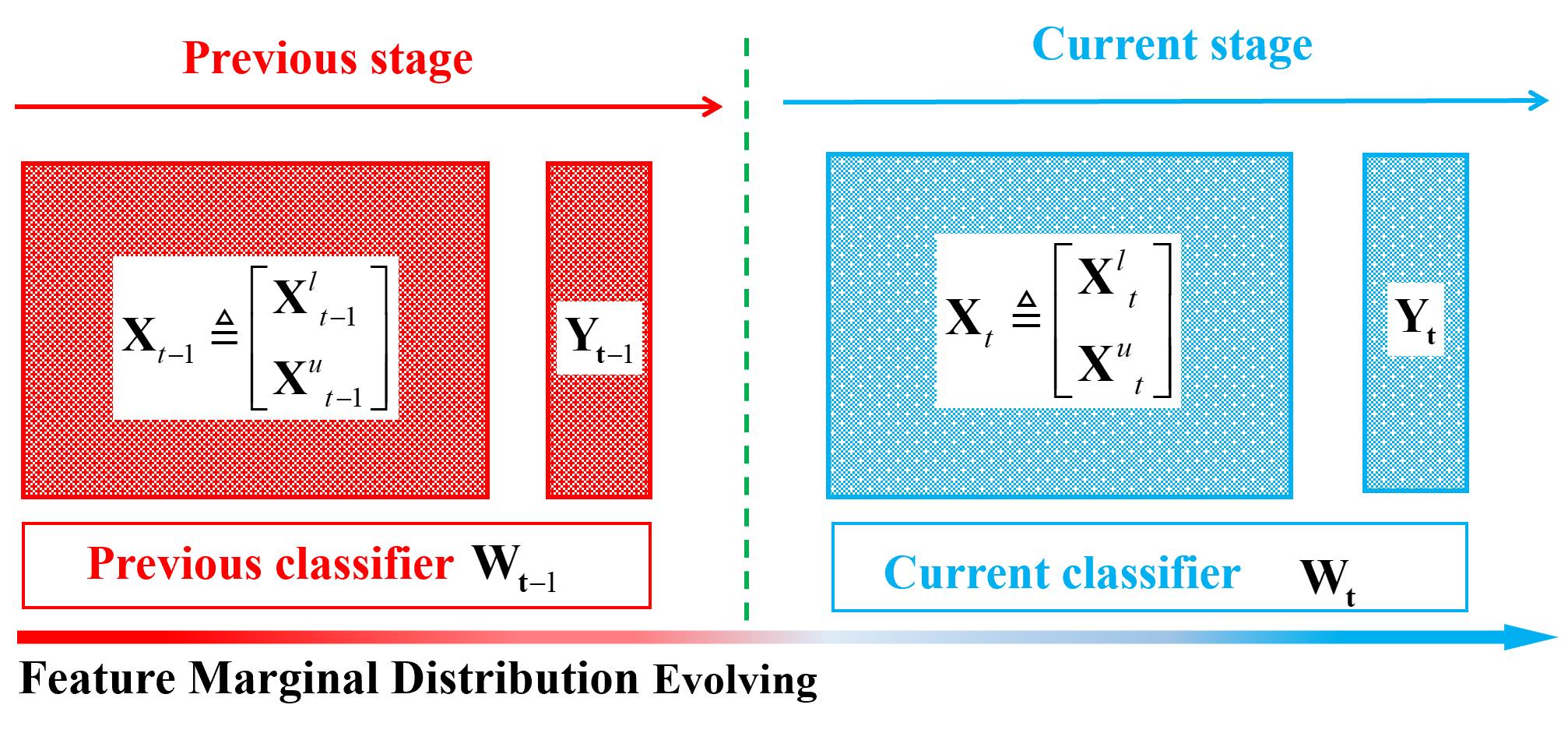}
	\end{center}
	\caption{Illustration for a local step in the evolving stage.}
	\label{figure2}
	\vskip -.2in
\end{figure}

According to the scenario of this article shown in Fig.\ref{figure1}, the distribution evolving process is divided into three stages, i.e., initiation, evolving, and ending stages, and the evolving stage will persist for a certain duration. Without loss of generality, we assume that the whole evolution proceeds in the same feature space $\mathcal{X}\subseteq \mathbb{R}^d$ and shares the same label space $\mathcal{Y}=\left\{{+1,-1} \right\}^C$ due to the invariance of the classification task. Suppose that we are provided with a evolving distribution $\mathcal{D}_t \in \mathcal{X} \times \mathcal{Y}, t\in[0,1]$. Here $\mathcal{D}_0=\mathcal{D}_S$ and $\mathcal{D}_1=\mathcal{D}_T$ correspond to the initiation and ending distributions respectively. To quantify the continually evolving nature of $\mathcal{D}_t$, we assume $\textbf{d}(\mathcal{D}_t ;\mathcal{D}_{t+1}) > 0$ for some distribution distance $\textbf{d}$. We further assume $\lim _{\Delta t \to 0}\textbf{d}(\mathcal{D}_t ;\mathcal{D}_{t+\Delta t}) = 0$ as the continuity of evolvement. During the evolution process, the data points come in batches sequentially $T=\{D_0, D_1, \cdots, D_T \}$. Due to the continuity of evolution, we assume that the data within each batch are independently and identically distributed. Accordingly, the model that tracks this evolving data distribution is defined as $f_t$, which starts at $f_0=f_S$ and ends at $f_1=f_T$. 

In initiation stage, also named preparation stage, the learner collects a sufficient labeled dataset $D_S = \{\mathbf{X}_S,\mathbf{Y}_S\}= \{(\bm{x}_{S_i},\bm{y}_{S_i})\}_{i=1}^{n}$ from source distribution $\mathcal{D}_S$. Source model $f_S:\mathcal{X}\mapsto \mathcal{Y}$ is well-trained on the source dataset $D_S$ with a small risk $R(f_S;\mathcal{D}_S)$. In the evolving stage, as shown in Fig. \ref{figure2}, we consider a local step in the evolution process. Following the pioneering work \cite{DBLP:journals/tkde/HouZZ21}, we assume that two consecutive batches of evolving data are available at each time in the data stream, i.e., the one-shot learning scenario. More specifically, each local step contains two batches of data, previous data $D_{t-1}$ and current data $D_t$. In particular, due to the expensive labeling cost, there are usually only a few collected labeled data, especially within a short period. At each moment $t$, we have $D_t=D_t^{l}\cup D_t^{u}$, where $D_t^{l}=\{\mathbf{X}_t^l,\mathbf{Y}_t^l\}=\{(\bm{x}_{t_i},\bm{y}_{t_i})\}_{i=1}^{l}$ is the labeled data and $D_t^{u}=\{\bm{x}_{t_i}\}_{i=l+1}^{u}$ is the unlabeled data. Here, we assume that the numbers of labeled and unlabeled points are the same in all evolving stages since we can only handle a fixed data volume at each moment. In summary, Table \ref{table 1} is the main notation.

\renewcommand\arraystretch{0.8}
\renewcommand\tabcolsep{40pt} %列间距
\begin{table}[h]
	\centering
	\caption{Main Notations and Corresponding Definitions.}
	\small
	\begin{tabular}{cc}
		\toprule[1pt]
		Notation & Definition \\ 
		\toprule[.75pt]
		$D_S$ & Data in initiation stage  \\
		$D_t$ & Data at evolution time $t$ \\
		$\mathbf{X}_S\in {\mathbb{R}^{{d}\times {n}}}$ & Feature matrix in initiation stage\\
		$\mathbf{X}_t\in {\mathbb{R}^{{d}\times {(l+u)}}}$ &Feature matrix at evolution time $t$\\
		$\mathbf{Y}_S\in {\mathbb{R}^{{C}\times {n}}}$ & Label matrix in initiation stage\\
		$\mathbf{Y}_t^l\in {\mathbb{R}^{{C}\times {l}}}$ & Label matrix at evolution time $t$\\
		$f_t\in {\mathcal{F}_t}$& Evolving model, $t\in [0,1]$\\
		\bottomrule[1pt]
	\end{tabular}
	\label{table 1}
	\vskip -.2in
\end{table}

\subsection{The DAL framework} 

For the data distribution evolving problem, our goal is to learn a well-generalized classifier $f_t$ to track the evolving data distribution. It achieves good ability by minimizing the expected risk over the evolving distribution. 
\begin{equation}\label{evolving expected risk}
R(f_t;\mathcal{D}_{(t)}) = \mathbb{E}_{(\bm{x},\bm{y})\sim\mathcal{D}_{(t)}}[\ell(f_t(\bm{x}),\bm{y})],
\end{equation}
where $\ell(\cdot): \mathbb{R}^C\mapsto\mathbb{R}$ is the loss function to measure the difference between vector form class affiliation prediction and the true label, the smaller the better. According to our assumption, the source learner has achieved a small generalization risk $R(f_S;\mathcal{D}_S)$ on the source distribution. Inspired by the idea of model reuse, the helpfulness of the model from a related task is stressed, we reuse the model of the former stage to assist the learning of the current stage model. The evolving data distribution tracing model $f_t$ commences with the source model $f_S$ and progressively adapts to the current task through inheritance and evolution. Besides, due to the limited labeled data available during the evolving stage, we incorporate manifold regularization into our approach to unveil latent data structural information and mitigate the challenges posed by limited labeled data, thereby enhancing both model performance and generalization ability. Consider a linear classifier $f_t(\bm{x}_{t_i})=\mathbf{W}_t^\top\bm{x}_{t_i}\in \mathbb{R}^C$ over the centralized instance $\bm{x}_i$ with model $\mathbf{W}_t\in \mathbb{R}^{d\times C}$, whose columns correspond to each class, we formulate the Distribution Adaptable Learning framework as following.
\begin{equation}\label{DAL framework}
\begin{aligned}
\mathop {\min }\limits_{\mathbf{W}_t \in \bm{\mathcal{W}}_t} \frac{1}{l}\sum\limits_{i = 1}^{l} {\ell\left( {f_t\left( {\bm{x}_{t_i}} \right),{y}_{t_i}} \right)}  + \alpha\Omega({\mathbf{W}_t},{\mathbf{W}_{t-1}}) +\beta \mathcal{R} ({\mathbf{X}_t},{\mathbf{W}_t}),
\end{aligned}
\end{equation}
where $\Omega({\mathbf{W}_t},{\mathbf{W}_{t-1}})$ refers to the regularization term with the help of pre-trained models, which is the key to model reuse. $\mathcal{R} (\mathbf{X}_t,{\mathbf{W}_t})$ refers to the regularization term with the help of unlabeled data, which is the key to semi-supervised learning. Subsequently, we will comprehensively examine two important components of the DAL framework, i.e., \textbf{Transport Model Reuse} and \textbf{Manifold Regularization}.

\subsection{Transport Model Reuse}

\begin{figure*}[!t]
	\setlength{\abovecaptionskip}{-.1cm}
	\setlength{\belowcaptionskip}{-.1cm}
	\begin{center}
		\includegraphics[width=13cm]{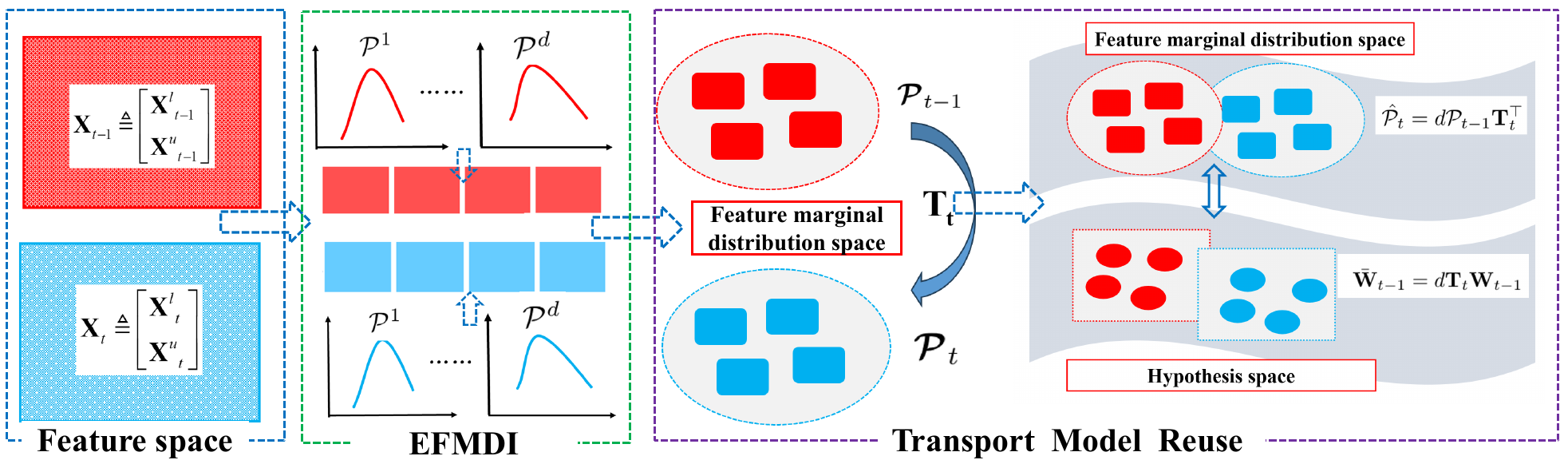}
	\end{center}
	\caption{Illustration of the EFMDI and \textbf{Transport Model Reuse}}
	\label{figure3}
\end{figure*}

Traditional model reuse is limited to the case of reusing a well-trained model within the same data distribution. However, the real-world environment is not stationary, the data distribution may constantly evolve. The violation of independent and identically distributed assumption poses an obstruction to model reuse. Facing this obstruction, inspired by the idea of optimal transport, we extend model reuse to the case of evolving data distributions by constructing a transformation between feature sets as well as model space, which enables model reuse across different data distributions. The primary idea of \textbf{Transport Model Reuse} is shown in Fig. \ref{figure3}.

\subsubsection{Optimal Transport Between Feature Domains}

Following the pioneering work \cite{DBLP:conf/icml/YeZ0Z18}, we consider that the variant data distribution across tasks can be substantially related, and model reuse on evolving data distribution should focus on the feature transformation between previous and current features domains. If each feature domain has a corresponding discrete marginal probability distribution, the map can be obtained by the coupling between their normalized marginal probability mass vectors $\bm{\mu}_1\in {\Delta _{d}}$ and $\bm{\mu}_2\in {\Delta _{d}}$. A transport plan is represented as a joint distribution, which is defined in the following domain, 
$$\mathcal{T}=\left\{\mathbf{T}\in \mathbb{R}_+^{d\times d}|\mathbf{T}\mathbf{1}_{d}=\bm{\mu}_1,\mathbf{T}\mathbf{1}_{d}=\bm{\mu}_2\right\},$$
and the entropy of $\mathbf{T}$ is defined as
$$E(\mathbf{T})=-\sum\limits_{ij} {T_{ij}(\log T_{ij}-1)}.$$
For the practicability and comprehensibility, we introduce a matrix $\mathbf{C}\in \mathbb{R}_+^{d \times d}$ to depict the feature variation relationship, i.e., the evolving cost of transforming features from former to current task. Optimal transport aims to find a transport matrix with the minimal cost, which is modeled as the following optimization problem:
\begin{equation}\label{OT}
	\hat{\mathbf{T}}= \mathop {\arg \min}\limits_{\mathbf{T}\in \mathcal{T}}  < \mathbf{T},\mathbf{C} >  + \alpha E(\mathbf{T}),
\end{equation}
where $\alpha$ is a trade-off parameter.

Eq. (\ref{OT}) is also the Kantorovitch formulation of the Optimal Transport (OT) problem \cite{Villani2014Optimal}, which aligns two distributions by the learned coupling $\mathbf{T}$. $\mathbf{T}$ demonstrates the process of constructing a transformation from one set to another, where the marginal probability mass of a feature is reallocated to similar features with minimal associated costs. This feature transformation can also be applied to the model space, i.e., coefficients in one model can be transported to another weighted by their feature similarity. For instance, in a straightforward scenario where we interchange the positions of features to construct a novel feature space, the cost matrix $\mathbf{C}$ will take on the form of a square permutation-like matrix that reveals the correspondence between said features. With uniform feature marginal, OT will output a permutation matrix that aligns two sets of features correctly \cite{DBLP:journals/pami/CourtyFTR17}. By applying this feature alignment technique to models, we can perfectly transform a "well-trained" classifier from a previous task to the current one. In general scenarios, it is also meaningful to transform models based on feature transportation plans since model coefficients for similar features usually have similar values. For example, if each feature represents a word and cost reflects their physical similarities, then predictor weights for "Trump" may be close to those of "Obama" \cite{DBLP:conf/icml/KusnerSKW15}. At this point, the problem of \textbf{Transport Model Reuse} becomes clear, we solely require the transport matrix $\mathbf{T}$ between the two feature domains to facilitate model reuse across different data distributions.

We propose the Distribution Adaptable Learning (DAL) framework to reuse the model from related former tasks even when the data distributions evolve. In detail, for the current task, the goal is to reuse a well-trained model $\mathbf{W}_{t-1}\in \mathbb{R}^{d\times C}$ from the related former task. The primary idea of DAL is to utilize the semantic map $\mathbf{T}\in \mathbb{R}_+^{d\times d}$ between two feature domains to link models by setting the prior $\bar{\mathbf{W}}_{t-1} = d\mathbf{T}\mathbf{W}_{t-1}$. Here, $d$ is in the transformation and scales the marginal probability. Based on the above discussion, we formulate the transport model reuse regularization $\Omega({\mathbf{W}_t},{\mathbf{W}_{t-1}})$ as follows.
$$\Omega({\mathbf{W}_t},{\mathbf{W}_{t-1}})=\left\| {{\mathbf{W}_t}-d\mathbf{T}\mathbf{W}_{t-1}} \right\|_F^2.$$

\subsubsection{Evolving Cost Matrix Calculation}

Looking back at Eq. (\ref{OT}), it is clear that the key to achieving Optimal Transport lies in the construction of the cost matrix $\mathbf{C}$, which fully characterizes the change of features and the relationship between evolving data distribution. Each element of $\mathbf{C}$ represents the disparity of the corresponding two features. Nevertheless, quantifying the disparity between two features poses a new challenge because the dimensions of the feature vectors are not consistent due to the different data sizes of consecutive batches. According to research \cite{DBLP:journals/pami/CourtyFTR17}, sample-based optimal transport has achieved satisfactory results in domain adaptation problems. On the contrary, feature-based optimal transport has rarely been studied, because it is unreasonable to measure the distance between two vectors with different dimensions from the perspective of vector space. 

The latent feature marginal distribution gives a new turn to the dilemma. However, the marginal distribution of features is unavailable directly. To make it possible for the optimal transport to perform transformation across diverse feature domains. We propose a novel strategy named Encoding Feature Marginal Distribution Information (EFMDI), which makes an effective representation for each feature marginal probability distribution. Based on EFMDI, it becomes easy to compute the evolving cost matrix $\mathbf{C}$. For instance, $\mathbf{C}$ can be computed as the pairwise (squared) Euclidean distance between corresponding feature marginal distribution encoding vectors. Benefiting from EFMDI, we can effectively depict feature specifications and facilitate the construction of feature relationships. Therefore, the dynamic data distribution can be more effectively monitored, especially in a nonstationary environment with evolving feature distribution. Here we introduce two classical methods to implement the EFMDI to adapt to different realistic requirements.

(1) \textbf{Encoding Feature Marginal Distribution Information with Kernel mean embedding (KME)} \cite{DBLP:journals/corr/MuandetFSS16}. KME describes the probability distribution concisely, without information loss and supports convenient operations like mean calculation. It maps a probability distribution $\mathcal{P}$ defined over $\mathcal{X}$ to a reproducing kernel Hilbert space (RKHS) as a mean function
$$\mu_k(\mathcal{P}) \triangleq \int_\mathcal{X} {k(\bm{x},\cdot)} \text{d}\mathcal{P}(\bm{x}),$$
where $k:\mathcal{X}\times \mathcal{X}\to \mathbb{R}$ is a symmetric and positive definite kernel function \cite{DBLP:books/lib/ScholkopfS02}, with associated RKHS $\mathcal{H}_k$ and feature map $\phi:\mathcal{X}\to\mathcal{H}_k$. The embedding $\mu_k(\mathcal{P})$ exists and belongs to $\mathcal{H}_k$, when $\mathbb{E}_{\bm{x}\sim\mathcal{P}}[\sqrt{k(\bm{x},\bm{x})}]<\infty$.
According to the investigation of \cite{DBLP:journals/jmlr/SriperumbudurFL11}, there will be no loss of information regarding the distribution $\mathcal{P}$, when equipped with characteristic kernels such as Gaussian kernel. Since the favorable properties of KME in representing distribution, it is potentially a nice choice for encoding marginal distribution information of features.

In practice, we can only observe a dataset $D = \{\bm{x}_n\}_{n=1}^{N}$ sampled from the underlying distribution $\mathcal{P}$. Thus, the empirical KME $\hat\mu_k(\mathcal{P})$ is calculated on the dataset to approximate the expected version $\mu_k(\mathcal{P})$ as
$$\hat\mu_k(\mathcal{P})=\frac{1}{N}\sum\limits_{n = 1}^N {k(\bm{x}_n,\cdot)}.$$
Based on this, encoding marginal distribution information of features can be implemented by performing KME on each feature in the feature set. Precisely, for the $\mu$-th feature $F^\mu$ within the feature set, we possess
$$\hat\mu_k(\mathcal{P}^\mu)=\frac{1}{N}\sum\limits_{n = 1}^N {k({x}_{n\mu},\cdot)},$$
where ${x}_{n\mu}$ is the $\mu$-th component, i.e., the $\mu$-th feature of $\bm{x}_n$. After obtaining the marginal distribution representation of each feature $F^\mu$, we can calculate the evolving cost matrix $\mathbf{C}$ for two consecutive batches of data at moment $t$
\begin{equation}\label{EC}
	C^t_{\mu\nu}=\left\| {\hat\mu^t_k(\mathcal{P}^\mu)-\hat\mu^{t-1}_k(\mathcal{P}^\nu)} \right\|^2_{\mathcal{H}_k}, \mu,\nu\in [d].
\end{equation}
The aforementioned equation appears to pose a computational challenge, thus, we propose the following transformation. First, we introduce the following two matrices
$$\mathbf{K}^{\mu\nu} = \left[ \begin{gathered}
	\mathbf{K}_{t,t}^{\mu\mu} \hfill \\
	\mathbf{K}_{t,t-1}^{\mu\nu} \hfill \\ 
\end{gathered}  \right.\left. \begin{gathered}
	\mathbf{K}_{t-1,t}^{\nu\mu} \hfill \\
	\mathbf{K}_{t-1,t-1}^{\nu\nu} \hfill \\ 
\end{gathered}  \right],\quad H_{ij} = \left\{ \begin{gathered}
	\frac{1}{n^2_2}\quad x_{i\mu},x_{j\mu} \in F_{t}^\mu,\\
	\frac{1}{{{n^2_1}}} \quad x_{i\nu},x_{j\nu} \in  F_{t-1}^\nu, \\
	-\frac{1}{{{n_1n_2}}}\enspace otherwise\\ 
\end{gathered}  \right.$$
After that, we convert Eq. (\ref{EC}) into the following easily solvable form
$$C^t_{\mu\nu}=\textrm{Trace}(\mathbf{K}^{\mu\nu}\mathbf{H}).$$

(2) \textbf{Encoding Feature Marginal Distribution Information with Kernel Density Estimation (KDE)} \cite{DBLP:journals/pami/DevroyeM85}. The KDE method, proposed by Parzen \cite{1962On}, is utilized for estimating the unknown density function in probability theory and falls under the category of nonparametric testing methods. KDE is very similar to a histogram, but it is smooth and continuous. In particular, given a dataset $D = \{\bm{x}_n\}_{n=1}^{N}$ sampled \text{i.i.d.} from the underlying distribution $\mathcal{P}$. For any feature $F^\mu$ within the feature set, we have \text{i.i.d.} sample points $\{\bm{x}_{n\mu}\}_{n=1}^{N}$, where $\mu \in [d]$. Denote the probability density function as $f$, and the KDE is given by the following expression.
$${\hat p^\mu_h}(x) = \frac{1}{{Nh}}\sum\limits_{n = 1}^N {k(\frac{{x - {x_{n\mu}}}}{h})}.$$
Here $h>0$ is a smoothing parameter called bandwidth, which determines the variance of the kernel function. $k(\cdot)$ is the kernel function that satisfies nonnegative, symmetry, integral is 1, and conforms to the probability density property. There are many kinds of kernel functions, uniform, triangular, biweight, triweight, Epanechnikov, normal, etc. Due to the convenient mathematical properties of the Gaussian kernel, it is also common to use $k(x)= \phi(x)$, where $\phi(x)$ is the standard normal probability density function. Then the KDE can be written as 
$${\hat p^\mu_h}(x) = \frac{1}{{Nh}}\sum\limits_{n = 1}^N {\phi(\frac{{x - {x_{n\mu}}}}{h})}.$$
Subsequently, a question that arises naturally is how to determine the optimal value of $h$. Considering the sample size $N$, if we select a value of $h$ that is excessively large, it will inevitably fail to satisfy the requirement for $h$ to tend towards 0. Conversely, if $h$ is chosen too small, there will be an insufficient number of data points available for estimating $p(x)$. This phenomenon exemplifies the tradeoff between bias and variance in nonparametric estimation. According to the research of Parzen \cite{1962On}, there is an $h$ that minimizes the mean square error between $\hat p_h(x)$ and $p(x)$ in theory, that is $h=c\times N^{-1/5}$. For the Gaussian distribution, $c$ is equal to 1.05 times the standard deviation.

After obtaining the marginal distribution representation ${\hat p^\mu_h}(x)$ of each feature $F^\mu$, we can calculate the evolution cost matrix $\mathbf{C}$ for two consecutive batches of data at moment $t$
\begin{equation}\label{EC1}
	C^t_{\mu\nu}=\int\left[{\hat p^\mu_h}(x)^{t}-{\hat p^\nu_h}(x)^{t-1} \right]^2\text{d}x, \mu,\nu\in [d].
\end{equation}

\subsection{Manifold Regularization}

Considering the costly nature of labeling, the evolution scenario presented in this paper typically involves a limited number of labeled data collected at each moment. However, training a model solely on such limited labeled data would overlook substantial valuable information inherent in unlabeled data. Drawing inspiration from the principles of semi-supervised learning \cite{DBLP:conf/cis/PiseK08}, the performance enhancement of DAL can be facilitated by simultaneously considering instance structural information and label correlations. On the whole, we assume that label correlations can be reflected by the topological relationship of instances. Specifically, a similarity graph is constructed to represent the topological relationship of instances, based on which the relationship between labels is depicted with certain assumptions. In this way, the instance structural information is transferred from the feature space to the label space. The technical details necessitate the incorporation of pertinent techniques from semi-supervised learning paradigms, such as label propagation (LP) \cite{DBLP:journals/tkde/ZhangZFLG21} and manifold learning \cite{DBLP:conf/aaai/HouGZ16}. 

Graph regularization, a representative manifold regularization, provides us the flexibility to incorporate our prior knowledge on some particular applications. When a set of unlabeled examples is available, the local invariance assumption guides Graph regularization to construct a similarity graph matrix ${\bm{\mathrm{G}}}$ based on the correlation between instances, thereby effectively reflecting the interrelationships among labels present in the training data. Hence, $\mathcal{R} (\mathbf{X}_t,{\mathbf{W}_t})$ can be formulated as $\textrm{Tr}({\mathbf{W}_t}^\top\mathbf{X}_t)\mathbf{G}({\mathbf{W}_t}^\top\mathbf{X}_t)^\top$, where the graph Laplacian matrix ${\bm{\mathrm{G}}} \in {{\bm{\mathbb{R}}}^{n \times n}}$ is calculated in the feature space. Specifically, given a set of examples $\left\{ {\bm{{x}}_i} \right\}_{i = 1}^n$, we can use a $k$-nearest neighbor graph to model the relationship between nearby data points. Define the corresponding weight matrix be $\bm{\mathrm {A}}$,
\begin{equation} \label{LP}
A_{i j}=\left\{\begin{array}{lll}
\exp(- \frac{\left\|{\bm{{x}}_i-\bm{{x}}_j} \right\|_{F}^{2}} {\sigma^{2}}), &\bm{{x}}_i \in {N_k}(\bm{{x}}_j)~\rm{and}~\bm{{x}}_\mathit{j} \in \mathit{{N_k}}(\bm{{x}}_\mathit{i}) \\
0, &  \text{ otherwise }
\end{array}\right.,
\end{equation}
where $\sigma$ is the variance of the Gaussian kernel and is usually estimated according to the average distance between sample points, ${{N_k}({\bm{{x}}_i})}$ is the set of $k$ neighbors of sample point $\bm{x}_i$. Then the Laplacian matrix is obtained by $\bm{\mathrm {G}}=\bm{\mathrm {D}} - \bm{\mathrm{A}}$, where $\bm{\mathrm {D}}$ is a diagonal matrix with diagonal elements ${{D}_{ii}}=\sum\nolimits_{j = 1}^n {{{A}_{ij}}}$. In short, a representative manifold regularization is suggested, the third part can be formulated as but not limited to the following form
\begin{equation}\label{SSR}
\mathcal{R} (\mathbf{X},{\mathbf{W}_t}) = 
\textrm{Tr}({\mathbf{W}_t}^\top\mathbf{X}_t)^\top\mathbf{G}({\mathbf{W}_t}^\top\mathbf{X}_t).
\end{equation}

The completion of each component elucidates the DAL model framework, rendering it comprehensive and coherent. Eq. (\ref{DAL framework}) can be presented in more detail as
\begin{equation}\label{DAL}
\begin{aligned}
\mathop {\min }\limits_{\mathbf{W}_t \in \bm{\mathcal{W}}_t} \frac{1}{l}\sum\limits_{i = 1}^{l} {\ell\left( {f\left( {\bm{x}_{t_i}} \right),{y}_{t_i}} \right)} + \alpha\left\| {{\mathbf{W}_t}-d\mathbf{T}\mathbf{W}_{t-1}} \right\|_F^2 +\beta \textrm{Tr}({\mathbf{W}_t}^\top\mathbf{X}_t)^\top\mathbf{G}({\mathbf{W}_t}^\top\mathbf{X}_t).
\end{aligned}
\end{equation}
Subsequently, we present a meticulous theoretical analysis of the DAL framework from the perspective of generalization theory to illustrate the effectiveness of DAL.

\section{Theoretical Analysis}

In this part, we conduct a detailed theoretical analysis of the DAL framework, including the generalization error bound of DAL within a local step and the generalization error bound associated with the classifier trajectory of the entire evolution. In addition, the improvement of \textbf{Transport Model Reuse} and \textbf{Manifold Regularization} on the generalization ability of the DAL model framework are further discussed. Before going into the details, we need the following lemmas. 
\begin{lemma}[Generalization Error Bound]
	\label{Lemma1}
	Let $\bm{\mathcal{L}}$ be the family of loss function associated to $\bm{\mathcal{F}}$, i.e., $\bm{\mathcal{L} }= \left\{ {{\bm{x}} \to \ell(f(\bm{x}),\bm{y}),f \in \bm{\mathcal{F}}} \right\}$. Suppose the loss function is $B$-bounded, then, for any $\delta $ > 0, with probability at least $1-\delta $ over a sample of size $m$, the following inequality
	holds for all $f \in \bm{\mathcal{F}}$:
	\begin{equation}\label{Generalization Error Bound}
	R_{\mathcal{D}}(f)\leq\hat R_{m}(f)+2{\hat{\mathfrak{R}}_m }(\bm{\mathcal{L}})+3B\sqrt{\frac{{\log (1/\delta )}}{2m}},
	\end{equation}
	where ${\hat{\mathfrak{R}}}(\bm{\mathcal{L}})$ is empirical Rademacher complexity of loss function class $\bm{\mathcal{L}}$ associated to $\bm{\mathcal{F}}$, which can be bounded by using the celebrated Talagrand`s lemma \cite{1976Probability}.
\end{lemma}

\begin{lemma}[Rademacher Vector Contraction Inequality \cite{DBLP:journals/corr/Maurer16}]
	\label{Lemma2}
	Let ${\mathcal{T}}$ be a class of real functions, and ${\bm{\mathcal{H}}} \subset {\bm{\mathcal{T}}} = {{\bm{\mathcal{T}}}_1} \times \cdots \times {{\bm{\mathcal{T}}}_C}$ be a $C$-valued function class. If $\Phi :\mathbb{R}^C \mapsto \mathbb{R}$ is a $L$-Lipschitz continuous function, the following inequality holds
	$${\hat {\mathfrak{R}}_{ m}}(\Phi  \circ {\bm{\mathcal{H}}}) \le \sqrt 2 L\sum\nolimits_{i = 1}^C {{\hat {\mathfrak{R}}_{m}}} ({{\bm{\mathcal{T}}}_i})$$.
\end{lemma}

\begin{lemma}[Kahane-Khintchine inequality \cite{1994On}]
	\label{Lemma3}
	For any vectors $\bm{a}_1,\cdots,\bm{a}_n$ in a Hilbert space and independent Rademacher random variables $\sigma_1,\cdots,\sigma_n$ we
	have
	$$\frac{1}{\sqrt{2}}\mathbb{E}\left\| \sum\limits_{i = 1}^n\sigma_i\bm{a}_i\right\|^2\leq\left(\mathbb{E}\left\| \sum\limits_{i = 1}^n\sigma_i\bm{a}_i\right\|\right)^2\leq\mathbb{E}\left\| \sum\limits_{i = 1}^n\sigma_i\bm{a}_i\right\|^2$$
\end{lemma}

\subsection{Generalization Ability of DAL}

First of all, we analyze the generalization ability of each local step of the DAL framework, a standard generalization bound of the model in Eq. (\ref{DAL}) is presented as follows.

\begin{theorem} \label{Theorem1}
	Considering the C-classification task with a linear model under data distribution evolving. At each moment of evolution, let $\bm{\mathcal{H}}_1,\cdots, \bm{\mathcal{H}_C}$ be function classes of DAL, then denote $$\bm{\mathcal{H}}=\mathop  \oplus _{j \in [C]} \bm{\mathcal{H}}_j = \left\{ {\bm{x} \mapsto \bm{h}(\bm{x})=[h_1(\bm{x}),\cdots,h_C(\bm{x})]^\top: h_j\in \bm{\mathcal{H}}_j } \right\}$$ as the family of hypothesis set with  $h_j:\bm{w}_j^\top\bm{x}\mapsto \mathbb{R}$. Denote $\bm{h}\in \bm{\mathcal{H}}$ as the hypothesis returned by the model trained with the data ${D}_{t}\sim\mathcal{D}_{t}$ at this moment. The sample ${D}_{t}={D}_l\cup {D}_u$, ${D}_l$ is labeled data and ${D}_u$ is unlabeled data. Suppose the loss function $\ell$ is $B$-bounded and $L$-Lipschitz w.r.t. Euclidean norm. Then, for any $\delta>0$, with probability at least $1-\delta$ the following inequalities holds for all $\bm{h}\in \bm{\mathcal{H}}$
	\begin{equation}\label{T1-1}
		R(\bm{h};\mathcal{D}_{t}) \leq \hat R_{l}(\bm{h})+2\sqrt 2 L\sum\nolimits_{j = 1}^C {{\hat {\mathfrak{R}}_{l}}} ({\bm{\mathcal{H}}_j}) + 3B \sqrt{\frac{{\log (1/\delta )}}{2l}}.
	\end{equation}
\end{theorem}
Here, $R(\bm{h};\mathcal{D}_{t})$ and $\hat R_{l}(\bm{h})$ are the generalization error and empirical error, ${{\hat {\mathfrak{R}}_{l}}} ({\bm{\mathcal{H}}_j})$ is the empirical Rademacher Complexity \cite{DBLP:conf/colt/BartlettM01} of hypothesis class $\bm{\mathcal{H}}_j, j\in[C]$. \\
\textbf{Proof of Theorem \ref{Theorem1}}~~~
The proof of Theorem \ref{Theorem1} will start from the standard Rademacher complexity bound for $\bm{\mathcal{H}}$. For any $\delta > 0$, with probability at least $1-\delta $ over a sample ${D}_{t}={D}_l\cup {D}_u$, the following inequality
holds for all $\bm{h}\in \bm{\mathcal{H}}$:
\begin{equation}\label{PT1-1}
	\begin{aligned}
		R(\bm{h};\mathcal{D}_{t}) &\mathop\leq\limits^{\textcircled{\small{1}}} \hat R_{l}(\bm{h})+2{\hat{\mathfrak{R}}}(\mathcal{L}\circ\bm{\mathcal{H}}) + 3B \sqrt{\frac{{\log (1/\delta )}}{2l}}\\
		&\mathop\leq\limits^{\textcircled{\small{2}}}\hat R_{l}(\bm{h})+2\sqrt 2 L\sum\nolimits_{j = 1}^C {{\hat {\mathfrak{R}}_{l}}} ({\bm{\mathcal{H}}_j}) + 3B \sqrt{\frac{{\log (1/\delta )}}{2l}}
	\end{aligned}.
\end{equation}
where the inequality ${\textcircled{\small{1}}}$ holds due to Lemma \ref{Lemma1}, a standard generalization error analysis based on the empirical Rademacher complexity argument. Then, the latter inequality ${\textcircled{\small{2}}}$ is obtained according to Lemma \ref{Lemma2}.

Reaching a step further, for ${{\hat {\mathfrak{R}}_{l}}} ({\bm{\mathcal{H}}_j})$, we can derive the subsequent bound.
\begin{theorem} \label{Theorem2}
	For the Empirical Rademacher Complexity of hypothesis class $\bm{\mathcal{H}}_j, j\in[C]$ in Theorem \ref{Theorem1}, the following inequalities holds
	\begin{equation}\label{T2-1}
		\frac{1}{{\sqrt[4]{2}}}\frac{U}{l} \leq \hat {\mathfrak{R}}_l({\bm{\mathcal{H}}}_j)\leq \frac{U}{l},
	\end{equation}
	where
	$$U^2=\tilde{\alpha}^{-1}\textrm{Tr}(\mathbf{K}_{ll})-\tilde{\beta}\textrm{Tr}\left(\mathbf{J}^\top(\mathbf{I}+\tilde{\beta}\mathbf{M})^{-1}\mathbf{J}\right)$$ with $\mathbf{I}$ the identity matrix, and $$\mathbf{J}=\tilde{\alpha}^{-1}\mathbf{G}^\top\mathbf{K}_l^\top,\enspace \mathbf{M}=\tilde{\alpha}^{-1}\mathbf{G}^\top\mathbf{K}\mathbf{G}.$$ $\tilde{\alpha}=\frac{C\alpha}{\hat R_l(\bar {\mathbf{W}}_{t-1})}$ and $\tilde{\beta}=\frac{C\beta}{\hat R_l(\bar{\mathbf{W}}_{t-1})}$ are the transformed model parameters. $\bar {\mathbf{W}}_{t-1}$ is the transformed hypothesis associated with the hypothesis returned by the model at the previous moment and $\hat R_l(\bar {\mathbf{W}}_{t-1})$ is the empirical error of $\bar {\mathbf{W}}_{t-1}$. 
\end{theorem}

As seen from Theorems \ref{Theorem1} and \ref{Theorem2}, we can get some observations about the proposed DAL framework. On one hand, Theorem \ref{Theorem1} indicates that, with the increase of labeled samples, the empirical error keeps approaching the generalization error. In the extreme case when the labeled data is infinite, $R(\bm{h};\mathcal{D}_{t})$=$\hat R_{l}(\bm{h})$, which is consistent with intuition. On the other hand, Theorem \ref{Theorem2} suggests that the empirical Rademacher Complexity of hypothesis class $\bm{\mathcal{H}}_j$ can be bounded within a range closely associated with the model. This observation provides valuable guidance for designing more robust models to achieve higher accuracy in identifying target hypotheses when labeled data is limited. \\
\textbf{Proof of Theorem \ref{Theorem2}}~~~
The techniques adopted to derive $\hat {\mathfrak{R}}_l({\bm{\mathcal{H}}}_j)$ is inspired by those used for analyzing co-RLS in \cite{DBLP:conf/icml/SindhwaniR08,DBLP:journals/jmlr/RosenbergB07}.
Before going into the details, we will transform the optimization problem and introduce some notation and preliminaries. Recalling that our original problem is to minimize $$Q({\mathbf{W}_t}):=\hat R_l({\mathbf{W}_t})+\alpha\left\| {{\mathbf{W}_t}-\mathbf{OT}({\mathbf{W}_{t-1}})} \right\|_F^2+\beta \textrm{Tr}({\mathbf{W}_t}^\top\mathbf{X}_{t})^\top\mathbf{L}({\mathbf{W}_t}^\top\mathbf{X}_{t}).$$ 
Let $\bar {\mathbf{W}}_{t-1}=\mathbf{OT}({\mathbf{W}_{t-1}})$ and $\hat {\mathbf{W}}= {\mathbf{W}_t}-\bar {\mathbf{W}}_{t-1}\in  {\hat{\bm{\mathcal{H}}}}=\mathop  \oplus _{j \in [C]} \hat{\bm{\mathcal{H}}}_j$, we transform the original optimization problem into the following minimization problem about $\hat {\mathbf{W}}$
\begin{equation}\label{PT2-1}
	\begin{aligned}
		Q(\hat {\mathbf{W}})&:=\hat R_l(\hat {\mathbf{W}}+\bar {\mathbf{W}}_{t-1})+\alpha\left\| \hat {\mathbf{W}} \right\|_F^2+\beta \textrm{Tr}((\hat {\mathbf{W}}+\bar {\mathbf{W}}_{t-1})^\top\mathbf{X}_{t})^\top\mathbf{L}((\hat {\mathbf{W}}+\bar {\mathbf{W}}_{t-1})^\top\mathbf{X}_{t}). \\
		&:=\hat R_l(\hat {\mathbf{W}}+\bar {\mathbf{W}}_{t-1})+\alpha\left\| \hat {\mathbf{W}} \right\|_F^2+\beta \textrm{Tr}(\hat {\mathbf{W}}^\top\mathbf{X}_{t})^\top\mathbf{L}(\hat {\mathbf{W}}^\top\mathbf{X}_{t})+\beta\textrm{Tr}(\bar {\mathbf{W}}_{t-1}^\top\mathbf{X}_{t})^\top\mathbf{L}(\bar {\mathbf{W}}_{t-1}^\top\mathbf{X}_{t}).
	\end{aligned}
\end{equation}
Then we obtain the equivalent optimization problem to minimize 
$$\bar Q(\hat {\mathbf{W}}):= \hat R_l(\hat {\mathbf{W}}+\bar {\mathbf{W}}_{t-1})+\alpha\left\| \hat {\mathbf{W}} \right\|_F^2+\beta \textrm{Tr}(\hat {\mathbf{W}}^\top\mathbf{X}_{t})^\top\mathbf{L}(\hat {\mathbf{W}}^\top\mathbf{X}_{t}).$$

Next, for each function class ${\hat{\bm{\mathcal{H}}}}_j$, we denote the reproducing kernels corresponding to ${\hat{\bm{\mathcal{H}}}}_j$ by $k:\bm{\mathcal{X}}\times \bm{\mathcal{X}}\to \mathbb{R}$. According to the Representer Theorem \cite{1994On}, it's convenient to introduce notation for the `span of the data' in the space ${\hat{\bm{\mathcal{H}}}}_j:=span\left\{k(\bm{x}_i,\cdot)\right\} $. According to the Representer Theorem \cite{1971Some,2008Semi}, it's clear that $\hat {h}_j^*\in \hat{\bm{\mathcal{H}}}_j$. That is, we can write the solution of $\bar Q(\hat {\mathbf{W}})$ as the following form
$$\hat{h}_j^*(\cdot)=\hat{\bm{w}}_j^*(\cdot)=\sum\nolimits_{i = 1}^{l + u} {\alpha_i^* k(\bm{x}_i,\cdot)}\in \hat{\bm{\mathcal H}}_j,j\in[C],$$
for $\bm \alpha^{*}=(\alpha_1^*,\alpha_2^*,\cdots, \alpha_{l+u}^*)^\top$. We'll denote an arbitrary element of $\hat{\bm{\mathcal H}}_j$ by $ \hat{\bm{w}}_j=\sum\nolimits_{i = 1}^{l + u} {\alpha_i k(\bm{x}_i,\cdot)}$.

Define the kernel matrices $K_{ij} = k(\bm{x}_i, \bm{x}_j)$, and partition them into blocks corresponding to labeled and unlabeled points:
$$\mathbf{K}=\left( \begin{gathered}
	\hfill \mathbf{K}_{ll}\\
	\hfill \mathbf{K}_{lu}^\top\\ 
\end{gathered}  \right.\left. \begin{gathered}
	\hfill \mathbf{K}_{lu}\\
	\hfill \mathbf{K}_{uu}\\ 
\end{gathered}  \right), \enspace \mathbf{K}_L=(\mathbf{K}_{ll}\enspace \mathbf{K}_{lu})$$
Then we can rewrite $\|\hat{\bm{w}}_j\|^2$ and $\hat{\bm{w}}_j^\top\mathbf{X}_{t}$ as
$$\|\hat{\bm{w}}_j\|^2=\bm {\alpha}^{\top}\mathbf{K}\bm{\alpha},\enspace \hat{\bm{w}}_j^\top\mathbf{X}_{t}=\mathbf{K}\bm{\alpha}.$$
After introducing the above notation and preliminaries, the proof of Theorem \ref{Theorem2} is carried out from the following steps. 

\begin{itemize}
	\item \textbf{Step 1: Bounding the DAL Hypothesis Class and Converting to Euclidean Space}
	
	First of all, we bound the DAL hypothesis class in combination with the regularizations and then convert from a supremum over the hypothesis space $\hat{\bm{\mathcal H}}$ to a supremum over a finite-dimensional Euclidean space that we can solve directly. Plugging in the trivial hypothesis $\hat {\mathbf{W}} \equiv \bm{0}$ gives the following upper bound:
	$$\mathop {\min }\limits_{\hat {\mathbf{W}}\in \hat{\bm{\mathcal{W}}}}\bar Q(\hat {\mathbf{W}}) \leq \bar Q(\bm{0})=\hat R_l(\bar {\mathbf{W}}_{t-1}).$$
	
	Since all terms of $\bar Q(\hat {\mathbf{W}})$ are nonnegative, we conclude that any $\hat {\mathbf{W}}^*$ minimizing $\bar Q(\hat {\mathbf{W}})$ is contained in
	\begin{equation}\label{PT2-2}
		\hat{\bm{\mathcal H}}:=\left\{\hat {\mathbf{W}}: \frac{\alpha}{\hat R_l(\bar {\mathbf{W}}_{t-1})}\left\| \hat {\mathbf{W}} \right\|_F^2+\frac{\beta}{\hat R_l(\bar {\mathbf{W}}_{t-1})} \textrm{Tr}(\hat {\mathbf{W}}^\top\mathbf{X}_{t})^\top\mathbf{L}(\hat {\mathbf{W}}^\top\mathbf{X}_{t})\leq 1 \right\},
	\end{equation}
	
	Expanding Eq. (\ref{PT2-2}) by columns, for each function class ${\hat{\bm{\mathcal{H}}}}_j$, we have
	\begin{equation}\label{PT2-3}
		{\hat{\bm{\mathcal{H}}}}_j:=\left\{\hat{\bm{w}}_j: \frac{C\alpha}{\hat R_l(\bar {\mathbf{W}}_{t-1})}\left\| \hat{\bm{w}}_j \right\|_F^2+\frac{C\beta}{\hat R_l(\bar {\mathbf{W}}_{t-1})} ({\hat{\bm{w}}_j}^\top\mathbf{X}_{t})^\top\mathbf{L}( {\hat{\bm{w}}_j}^\top\mathbf{X}_{t})\leq 1 \right\}.
	\end{equation}
	At each moment of evolution, the final hypothesis for the DAL algorithm is chosen from the class $$\bm{\mathcal H}:=\left\{\bm{x}\mapsto  {\mathbf{W}}_{t}^\top\bm{x}=(\hat {\mathbf{W}}+\bar {\mathbf{W}}_{t-1})^\top\bm{x}: \hat {\mathbf{W}}\in \hat{\bm{\mathcal H}} \right\}.$$
	Next, we convert a supremum over the hypothesis space $\bm{\mathcal{H}}$ to a supremum over a finite-dimensional Euclidean space that we can solve directly. According to the definition of empirical Rademacher Complexity, we have 
	$$\hat {\mathfrak{R}}_l({\bm{\mathcal{H}}})={\mathbb{E}_{\bm\sigma} }\left[ {\mathop {\sup }\limits_{\mathbf{W}_t \in {\bm{\mathcal{H}}}}\frac{1}{l} \sum\limits_{i = 1}^l {{\bm\sigma _i} {\mathbf{W}}_{t}^\top{\bm{x}_i}} } \right]={\mathbb{E}_{\bm\sigma} }\left[ {\mathop {\sup }\limits_{\hat {\mathbf{W}}\in \hat{\bm{\mathcal H}}} \frac{1}{l} \sum\limits_{i = 1}^l {{\bm\sigma _i}(\hat {\mathbf{W}}+\bar {\mathbf{W}}_{t-1})^\top{\bm{x}_i}} } \right]=\hat {\mathfrak{R}}_l(\hat{\bm{\mathcal H}}).
	$$
	Similarly, $\hat {\mathfrak{R}}_l({\bm{\mathcal{H}}}_j)=\hat {\mathfrak{R}}_l(\hat{\bm{\mathcal H}}_j)$, we focus on the bounded hypothesis space $\hat{\bm{\mathcal H}}_j$
	\begin{equation}\label{Prodf T-3}
		\hat {\mathfrak{R}}_l(\hat{\bm{\mathcal H}}_j)=\frac{1}{l}{\mathbb{E}_{\bm\sigma} } \mathop {\sup }\left\{\sum\limits_{i = 1}^l {{\sigma _i} {\hat{\bm{w}}_j}^\top{\bm{x}_i}}:{\hat{\bm{w}}_j}\in \hat{\bm{\mathcal H}}_j\right\} 
	\end{equation}
	As mentioned above $\|\hat{\bm{w}}_j\|^2=\bm {\alpha}^{\top}\mathbf{K}\bm{\alpha},\enspace \hat{\bm{w}}_j^\top\mathbf{X}_{t}=\mathbf{K}\bm{\alpha}$, let $\tilde{\alpha}=\frac{C\alpha}{\hat R_l(\bar{\mathbf{W}}_{t-1})}$ and $\tilde{\beta}=\frac{C\beta}{\hat R_l(\bar{\mathbf{W}}_{t-1})}$, we can rewrite the hypothesis class $\hat{\bm{\mathcal H}}_j$ as
	$$\hat{\bm{\mathcal H}}_j:=\left\{\hat{\bm{w}}_j: \tilde\alpha\bm {\alpha}^{\top}\mathbf{K}\bm{\alpha}+\tilde\beta\bm {\alpha}^{\top}\mathbf{K}\mathbf{L}\mathbf{K}\bm{\alpha}\leq 1 \right\}.$$
	And $\hat {\mathfrak{R}}_l(\hat{\bm{\mathcal H}}_j)$ can be written as
	\begin{equation}\label{Prodf T-4}
		\hat {\mathfrak{R}}_l(\hat{\bm{\mathcal H}}_j)=\frac{1}{l}{\mathbb{E}_{\bm\sigma} }\left[\mathop {\sup }\limits_{\bm\alpha}\left\{{\bm\sigma}^\top {\mathbf{K}_L}{\bm\alpha}\quad\text{s.t.} \enspace\tilde\alpha\bm {\alpha}^{\top}\mathbf{K}\bm{\alpha}+\tilde\beta\bm {\alpha}^{\top}\mathbf{K}\mathbf{L}\mathbf{K}\bm{\alpha}\leq 1\right\}\right].
	\end{equation}
	
	\item \textbf{Step 2: Evaluating the Supremum and Bounding it above and below}
	
	Secondly, we use the Kahane-Khintchine inequality to bound the expectation over $\bm{\sigma}$ above and below by expectations that we can compute explicitly. For a symmetric positive definite (spd) matrix $\mathbf{P}$, it's easy to show that $$\mathop {\sup }\limits_{\bm{\alpha}:{\bm{\alpha}}^\top \mathbf{P}\bm{\alpha}\leq1}\bm{v}^\top\bm{\alpha}=\left\|\mathbf{P}^{-1/2}\bm{v}\right\|.$$ 
	Note that each entry of the column vector ${\mathbf{K}}{\bm\alpha}$ is an
	inner product between ${\bm\alpha}$ and a row of ${\mathbf{K}}$. Thus if ${\bm\alpha _\parallel }=Proj_{space\mathbf{K}}$, i.e. the projection of $\bm\alpha$ onto each row of $\mathbf{K}$, then ${\mathbf{K}}{\bm\alpha}={\mathbf{K}}{\bm\alpha _\parallel}$ and that the quadratic form ${\bm{\alpha}}^{\top}(\tilde\alpha \mathbf{K}+\tilde\beta\mathbf{K}\mathbf{L}\mathbf{K}){\bm{\alpha}}$ is unchanged when we replace ${\bm\alpha}$ by $\bm\alpha _\parallel$. Thus the supremum can be rewritten as
	$$\hat {\mathfrak{R}}_l(\hat{\bm{\mathcal H}}_j)=\frac{1}{l}{\mathbb{E}_{\bm\sigma} }\left[\mathop {\sup }\limits_{\bm\alpha _\parallel}\left\{{\bm\sigma }^\top {\mathbf{K}_L}{\bm\alpha _\parallel}\quad\text{s.t.} \enspace{\bm\alpha _\parallel}^\top(\tilde\alpha \mathbf{K}+\tilde\beta\mathbf{K}\mathbf{L}\mathbf{K}){\bm\alpha _\parallel}\leq 1\right\}\right].$$
	Changing to eigenbases cleans up this expression and clears the way for substantial simplifications in later sections. Diagonalize the positive semi-definite (PSD) kernel matrices to get orthonormal bases for the column spaces of $\mathbf{K}$:
	$\mathbf{V}^\top\mathbf{K}\mathbf{V}=\bm\Sigma_\mathbf{K}$, where $\bm\Sigma_\mathbf{K}$ is a diagonal matrix containing the nonzero eigenvalues, and the columns of $\mathbf{V}$ is bases for the column spaces of $\mathbf{K}$. Now, we introduce $\bm{a}$ such that ${\bm\alpha _\parallel}=\mathbf{V}\bm{a}$. Applying this change of variables to the quadratic form, we get 
	$${\bm\alpha _\parallel}^\top(\tilde\alpha \mathbf{K}+\tilde\beta\mathbf{K}\mathbf{L}\mathbf{K}){\bm\alpha _\parallel}=\bm{a}^\top\mathbf{V}^\top(\tilde\alpha \mathbf{K}+\tilde\beta\mathbf{K}\mathbf{L}\mathbf{K})\mathbf{V}\bm{a}=\bm{a}^\top(\tilde\alpha \bm\Sigma_\mathbf{K}+\tilde\beta\mathbf{R}\mathbf{R}^\top)\bm{a}$$
	with $\mathbf{R}=\mathbf{V}^\top\mathbf{K}\mathbf{G}$ and $\mathbf{L}=\mathbf{G}\mathbf{G}^\top$. 
	
	Let $\mathbf{T}=\tilde\alpha \bm\Sigma_\mathbf{K}+\tilde\beta\mathbf{R}\mathbf{R}^\top$, $\mathbf{T}$ is spd, since it's the sum of the spd diagonal matrix $\bm\Sigma_\mathbf{K}$ and the spd matrix $\mathbf{R}\mathbf{R}^\top$. We can now write
	$$\hat {\mathfrak{R}}_l(\hat{\bm{\mathcal H}}_j)=\frac{1}{l}{\mathbb{E}_{\bm\sigma} }\left[\mathop {\sup }\limits_{\bm{a}}\left\{{\bm\sigma}^\top {\mathbf{K}_L}{\mathbf{V}}{\bm{a}}\quad\text{s.t.} \enspace{\bm{a}}^\top\mathbf{T}{\bm{a}}\leq 1\right\}\right].$$
	Since $\mathbf{T}$ is spd, we can evaluate the supremum as described above to get
	$$\hat {\mathfrak{R}}_l(\hat{\bm{\mathcal H}}_j)=\frac{1}{l}{\mathbb{E}_{\bm\sigma} }\left\|\mathbf{T}^{-1/2}\mathbf{V}^\top{\mathbf{K}_L}^\top{\bm\sigma}\right\|.$$
	Taking the columns of $\mathbf{T}^{-1/2}\mathbf{V}^\top{\mathbf{K}_L}^\top$ to be the $a_i$'s, we can apply Lemma \ref{Lemma3} to our expression for $\hat {\mathfrak{R}}_l(\hat{\bm{\mathcal H}}_j)$ to get 
	$$\frac{1}{{\sqrt[4]{2}}}\frac{U}{l} \leq \hat {\mathfrak{R}}_l(\hat{\bm{\mathcal H}}_j)\leq \frac{U}{l}$$
	where 
	$$U^2:={\mathbb{E}_{\bm\sigma} }\left\|\mathbf{T}^{-1/2}\mathbf{V}^\top{\mathbf{K}_L}^\top{\bm\sigma}\right\|^2={\mathbb{E}_{\bm\sigma} }\textrm{Tr}\left[{\mathbf{K}_L}\mathbf{V}\mathbf{T}^{-1}\mathbf{V}^\top{\mathbf{K}_L}^\top{\bm\sigma}{\bm\sigma}^\top\right]\mathop=\limits^{\textcircled{\small{1}}}\textrm{Tr}\left[{\mathbf{K}_L}\mathbf{V}\mathbf{T}^{-1}\mathbf{V}^\top{\mathbf{K}_L}^\top\right]$$
	Here ${\textcircled{\small{1}}}$ holds due to $\mathbb{E}{\bm\sigma}{\bm\sigma}^\top$ is the identity matrix.
	
	\item \textbf{Step 3: Writing the Expression in terms of the Original Kernel Matrices}
	
	Finally, with some matrix algebra, we can write $\hat {\mathfrak{R}}_l(\hat{\bm{\mathcal H}}_j)$ in terms of blocks of the original kernel matrices. It will be helpful to divide $\mathbf{V}$ into labeled and unlabeled parts, we have 
	$$\mathbf{K}=\left( \begin{gathered}
		\hfill \mathbf{K}_{ll}\\
		\hfill \mathbf{K}_{lu}\\ 
	\end{gathered}  \right.\left. \begin{gathered}
		\hfill \mathbf{K}_{ul}\\
		\hfill \mathbf{K}_{uu}\\ 
	\end{gathered}  \right)=\left( \begin{gathered}
		\mathbf{V}_l \hfill \\
		\mathbf{V}_u \hfill \\ 
	\end{gathered}  \right)\bm\Sigma_\mathbf{K}\left(\mathbf{V}_l^\top \enspace\mathbf{V}_u^\top \right).$$
	Rearranging the diagonalization, we also have
	$$\mathbf{V}^\top\left( \begin{gathered}
		\hfill \mathbf{K}_{ll}\\
		\hfill \mathbf{K}_{lu}\\ 
	\end{gathered}  \right.\left. \begin{gathered}
		\hfill \mathbf{K}_{ul}\\
		\hfill \mathbf{K}_{uu}\\ 
	\end{gathered}  \right)=\bm\Sigma_\mathbf{K}\left(\mathbf{V}_l^\top \enspace\mathbf{V}_u^\top \right).$$
	By equating blocks in these four matrix equations, we attain all the substitutions we need to write $U^2$ in terms of the original kernel submatrices $\mathbf{K}_{ll}$ and $\mathbf{K}_{lu}$. 
	$$\mathbf{V}^\top{\mathbf{K}_L}^\top=\bm\Sigma_\mathbf{K}\mathbf{V}_l^\top,\quad \mathbf{R}=\bm\Sigma_\mathbf{K}\mathbf{V}^\top\mathbf{G}.$$
	We now work on the $\mathbf{T}^{-1}$ factor in our expression. According to the Sherman-Morrison-Woodbury formula \cite{1996Matrix}, we expand $\mathbf{T}^{-1}=(\tilde\alpha \bm\Sigma_\mathbf{K}+\tilde\beta\mathbf{R}\mathbf{R}^\top)^{-1}$ as
	$$\begin{aligned}\mathbf{T}^{-1}=&\tilde\alpha^{-1}\bm\Sigma_\mathbf{K}^{-1}-\tilde\beta^{-1}\tilde\alpha^{-2}\bm\Sigma_\mathbf{K}^{-1}\bm\Sigma_\mathbf{K}\mathbf{V}^\top\mathbf{G}(\mathbf{I}+\tilde\beta^{-1}\tilde\alpha^{-1}\mathbf{G}^\top\mathbf{V}\bm\Sigma_\mathbf{K}\bm\Sigma_\mathbf{K}^{-1}\bm\Sigma_\mathbf{K}\mathbf{V}^\top\mathbf{G})^{-1}\mathbf{G}^\top\mathbf{V}\bm\Sigma_\mathbf{K}\bm\Sigma_\mathbf{K}^{-1}\\
		=&\tilde\alpha^{-1}\bm\Sigma_\mathbf{K}^{-1}-\tilde\beta^{-1}\tilde\alpha^{-2}\mathbf{V}^\top\mathbf{G}(\mathbf{I}+\tilde\beta^{-1}\tilde\alpha^{-1}\mathbf{G}^\top\mathbf{K}\mathbf{G})^{-1}\mathbf{G}^\top\mathbf{V}\end{aligned}.$$
	Since $\bm\Sigma_\mathbf{K}$ and $\mathbf{I}+\tilde\beta^{-1}\tilde\alpha^{-1}\mathbf{G}^\top\mathbf{K}\mathbf{G}$ are spd, the inverses exist and the expansion is justified. Substituting this expansion into the expression for $U^2$, we get
	$$\begin{aligned}U^2=&\textrm{Tr}\left[\tilde\alpha^{-1}\mathbf{V}_l\bm\Sigma_\mathbf{K}\bm\Sigma_\mathbf{K}^{-1}\bm\Sigma_\mathbf{K}\mathbf{V}_l^\top-\tilde\beta^{-1}\tilde\alpha^{-2}\mathbf{V}_l\bm\Sigma_\mathbf{K}\mathbf{V}^\top\mathbf{G}(\mathbf{I}+\tilde\beta^{-1}\tilde\alpha^{-1}\mathbf{G}^\top\mathbf{K}\mathbf{G})^{-1}\mathbf{G}^\top\mathbf{V}\bm\Sigma_\mathbf{K}\mathbf{V}_l^\top\right]\\
		=&\textrm{Tr}\left[\tilde\alpha^{-1}\mathbf{K}_{ll}-\tilde\beta^{-1}\tilde\alpha^{-2}{\mathbf{K}_L}\mathbf{G}(\mathbf{I}+\tilde\beta^{-1}\tilde\alpha^{-1}\mathbf{G}^\top\mathbf{K}\mathbf{G})^{-1}\mathbf{G}^\top{\mathbf{K}_L}^\top\right]\end{aligned}.$$
	Let $\mathbf{J}=\tilde{\alpha}^{-1}\mathbf{G}^\top\mathbf{K}_l^\top$, $\mathbf{M}=\tilde{\alpha}^{-1}\mathbf{G}^\top\mathbf{K}\mathbf{G}$, we get $U^2=\tilde{\alpha}^{-1}\textrm{Tr}(\mathbf{K_ll})-\tilde{\beta}\textrm{Tr}\left(\mathbf{J}^\top(\mathbf{I}+\tilde{\beta}\mathbf{M})\mathbf{J}\right)$, Theorem \ref{Theorem2} is proved.
\end{itemize}

Right after that, we move into another issue of interest, which is the generalization ability of the evolving distribution tracing model $\left\{{\bm{w}}_j(t)\right\}_{t\in[0,1]}$ throughout the evolution. The generalization bound is presented as follows.

\begin{theorem} \label{Theorem3}
	Considering the distribution evolving classification task with a linear model. At each moment of evolution, we denote ${\mathbf{W}}_t=[{\bm{w}}_j(t)], j\in[C]$ as the weight matrix corresponding to hypothesis $\bm{h}$. Suppose the loss function $\ell$ is $B$-bounded and $L$-Lipschitz w.r.t. Euclidean norm. Throughout the process of evolving data distribution, denote the weight trajectory as $\left\{{\bm{w}}_j(t)\right\}_{t\in[0,1]}$ for each partial hypothesis $\bm{h}_j$. Given a sequence $0=t_0\leq t_1\leq t_2,\cdots,\leq t_K\leq1$, the weight trajectory traces the evolving data distribution. Then, for any $\delta>0$, with probability at least $1-\delta $ the following inequalities holds 
	\begin{equation}\label{Tm3}
		\begin{aligned}
			\frac{1}{K}\sum\limits_{t=1}^K\left(R(\mathbf{W}_t;\mathcal{D}_{t})-\hat R_l(\mathbf{W}_t)\right)\leq\frac{1}{K}2\sqrt 2 L\sum\limits_{j = 1}^C\sum\limits_{k=1}^K \Delta t_k\mathop\mathbb{E}\limits_{{D}_{t}\sim\mathcal{D}_{t}}\left\|{\bm{w}}_j(t_k)'\right\|+ B\sqrt{\frac{{\log (1/\delta )}}{2l}},
		\end{aligned}
	\end{equation}
	where $\Delta t_k=t_k-t_{k-1}$,  ${\bm{w}}_j(t_k)'=\left\|\frac{\Delta{\bm{w}}_j(t_k)}{\Delta t_k}\right\|$ and $\Delta{\bm{w}}_j(t_k)={\bm{w}}_j(t_k)-{\bm{w}}_j(t_{k-1})$. As $\Delta t_k\to 0$, $$\sum\limits_{k=1}^K \Delta t_k\mathop\mathbb{E}\limits_{{D}_{t}\sim\mathcal{D}_{t}}\left\|{\bm{w}}_j(t_k)'\right\|\to \int_0^1 {\left\|\dot{\bm{w}}_j(t)\right\|}\enspace\text{d}t=\int_0^1{\sqrt {\left\langle\dot{\bm{w}}_j(t),\bm{g}(\bm{w}_j(t))\dot{\bm{w}}_j(t)\right\rangle}}\enspace\text{d}t,$$
	which is the length of the trajectory on the statistical manifold with inputs drawn from the interpolated distribution at each instant. 
\end{theorem}

This result is a crisp theoretical characterization of the intuitive idea that if one finds a weight trajectory that transfers from the source to the target task while obtaining a generalization error bound associated with the hypothesis space at each time instant, then the generalization ability of the tracking model is positively correlated with the length of the trajectory throughout the evolution. \\
\textbf{Proof of Theorem \ref{Theorem3}}~~~
The proof of Theorem \ref{Theorem3} will start from the manifold of probability distributions. Consider a manifold $\mathcal{M}= \left\{p_{\bm{w}}(\bm{z}) : \bm{w} \in \mathbb{R}^d\right\}$ of probability distributions. Weights $\bm{w}$ play the role of a coordinate system of $\mathcal{M}$. Information Geometry \cite{DBLP:conf/gsi/Amari13} studies invariant geometrical structures on such manifolds. For two
points $p_{\bm{w}}, p_{\bm{w}'}\in \mathcal{M}$, we can use the Euclidean divergence 
$$EU[p_{\bm{w}},p_{\bm{w}'}]=\int\frac{1}{2} (p_{\bm{w}}(\bm{z})-p_{\bm{w}'}(\bm{z}))^2\enspace\text{d}\bm{z},$$ 
to obtain a Riemannian structure on $\mathcal{M}$. For Euclidean divergence 
$EU[p_{\bm{w}},p_{\bm{w}'}]=EU[{\bm{w}},\bm{w}']$, which is a differentiable function of $\bm{w}$ and $\bm{w}'$. When $p_{\bm{w}}$ and $p_{\bm{w}+\text{d}\bm{w}}$ are sufficiently close, by denoting their coordinates by $\bm{w}$ and ${\bm{w}+\text{d}\bm{w}}$, the Taylor expansion of $EU$ is written as 
$$EU[{\bm{w}},{\bm{w}+\text{d}\bm{w}}]=\frac{1}{2}\sum\limits_{i,j=1}^dg_{ij}(\bm{w})\text{d}w_i\text{d}w_j+o(|\text{d}\bm{w}|^3).$$ 
This allows the infinitesimal distance $\text{d}s$ on the manifold to be written as
$$\text{d}s^2=2EU[{\bm{w}},{\bm{w}+\text{d}\bm{w}}]=\sum(\text{d}{w}_i)^2=\sum\limits_{i,j=1}^dg_{ij}({\bm{w}})\text{d}w_i\text{d}w_j,$$
$g_{ij}({\bm{w}})$ are elements of the Fisher Information Matrix (FIM) $\bm{g}$, which is the Hessian of the Euclidean divergence. As for Euclidean divergence, the FIM is an identity matrix. 

Next, we introduce the definition of Fisher-Rao distance. 

\begin{definition}[Fisher-Rao distance \cite{2021Information}] \label{definition3}
	Given a continuously differentiable curve $\left\{{\bm{w}}(\tau)\right\}_{\tau\in[0,1]}$ on
	the manifold $\mathcal{M}$ we can compute its length by integrating
	the infinitesimal distance $|\text{d}s|$ along it. The shortest length
	curve between two points $\bm{w}, \bm{w}'\in \mathcal{M}$ induces a metric on $\mathcal{M}$ known as the Fisher-Rao distance
	\begin{equation}\label{PT3-1}
		d_{FR}(\bm{w},\bm{w}')=\mathop{\min}\limits_{\begin{subarray}{l} \bm{w}:\bm{w}(0)=\bm{w}\\
				\enspace\bm{w}(1)=\bm{w}'\end{subarray} }\int_0^1{\sqrt{\left\langle\dot{\bm{w}}(\tau), \bm{g}({\bm{w}}(\tau))\dot{\bm{w}}(\tau)\right\rangle}}\enspace\text{d}\tau.
	\end{equation}
	The shortest paths on a Riemannian manifold are geodesics, i.e., they are locally ``straight lines''.
\end{definition}
Since the FIM is an identity matrix for Euclidean divergence. Observe that   
$${\left\langle\dot{\bm{w}}_j(t), \bm{g}({\bm{w}}_j(t))\dot{\bm{w}}_j(t)\right\rangle}={\left\langle\dot{\bm{w}}_j(t), \dot{\bm{w}}_j(t)\right\rangle}=\left\| \dot{\bm{w}}_j(t) \right\|^2$$

On the other hand, for moment $t$ let $\bm\Omega_t:=\left\{{\bm{w}}_j|\left\|{\bm{w}}_j-{\bm{w}}_j(t)\right\|\leq \left\|{\bm{w}}_j(t+\text{d}t)-{\bm{w}}_j(t)\right\| \right\}$ be a compact neighborhood of ${\bm{w}}_j(t)$ in weights space. The Rademacher complexity of $\Omega_t$ is upper bounded as following
\begin{equation}\label{PT3-2}
	\begin{aligned}
		{\mathfrak{R}}_l(\bm\Omega_t)&=\mathop\mathbb{E}\limits_{{D}_{t}\sim\mathcal{D}_{t}}{\mathbb{E}_{\bm\sigma} } \mathop {\sup\limits_{{{\bm{w}}_j}\in {\bm\Omega_t}} }\left[\frac{1}{l}\sum\limits_{i = 1}^l {{\sigma _i} {{\bm{w}}_j}^\top{\bm{x}_i}}\right]\\
		&=\mathop\mathbb{E}\limits_{{D}_{t}\sim\mathcal{D}_{t}}{\mathbb{E}_{\bm\sigma} } \mathop {\sup\limits_{{{\bm{w}}_j}\in {\bm\Omega_t}} }\left[\frac{1}{l}\sum\limits_{i = 1}^l {{\sigma _i} {{\bm{w}}_j}(t)^\top{\bm{x}_i}}+{{\sigma _i} {{\bm{w}}_j}^\top{\bm{x}_i}}-{{\sigma _i} {{\bm{w}}_j}(t)^\top{\bm{x}_i}}\right]\\
		&\leq\mathop\mathbb{E}\limits_{{D}_{t}\sim\mathcal{D}_{t}}{\mathbb{E}_{\bm\sigma} } \left[\frac{1}{l}\sum\limits_{i = 1}^l {{\sigma _i} {{\bm{w}}_j}(t)^\top{\bm{x}_i}}+\mathop {\sup\limits_{{{\bm{w}}_j}\in {\bm\Omega_t}} }\sum\limits_{i = 1}^l\frac{1}{l}{|{{\bm{w}}_j}^\top{\bm{x}_i}}-{{\sigma _i} {{\bm{w}}_j}(t)^\top{\bm{x}_i}}|\right]\\
		&=0+\mathop\mathbb{E}\limits_{{D}_{t}\sim\mathcal{D}_{t}}\left[\mathop {\sup\limits_{{{\bm{w}}_j}\in {\bm\Omega_t}} }\frac{1}{l}{|{{\bm{w}}_j}^\top{\bm{x}_i}}-{ {{\bm{w}}_j}(t)^\top{\bm{x}_i}}|\right]\\
		&\leq\mathop\mathbb{E}\limits_{{D}_{t}\sim\mathcal{D}_{t}}\left[\mathop{\sup\limits_{{{\bm{w}}_j}\in {\bm\Omega_t}} }\frac{1}{l}{\left\|{{\bm{w}}_j-{{\bm{w}}_j}(t)}\right\|\left\|{\bm{x}_i}\right\|}\right]\xrightarrow{\textcircled{\small{1}}}\mathop\mathbb{E}\limits_{\bm{x}\sim\mathcal{D}_{t}}\left[\mathop {\sup\limits_{{{\bm{w}}_j}\in {\bm\Omega_t}} }{\left\|{{\bm{w}}_j-{{\bm{w}}_j}(t)}\right\|\left\|{\bm{x}}\right\|}\right]\\
	\end{aligned},
\end{equation}
As $l$ goes to infinity. ${\textcircled{\small{1}}}$ holds due to compactness of $\bm\Omega_t$.  

Without loss of generality, assuming that $\left\|{\bm{x}}\right\|\leq1$, Given a trajectory of the weight $\left\{{\bm{w}}_j(t)\right\}_{t\in[0,1]}$ and a sequence $0=t_0\leq t_1\leq t_2,\cdots,\leq t_K\leq1$, we have 
$$\sum_{k=1}^K{\mathfrak{R}}_l(\bm\Omega_{t_k})\leq\sum_{k=1}^K\left\|{\bm{w}}_j(t_k)-{\bm{w}}_j(t_{k-1})\right\|=\sum_{k=1}^K\Delta t_k \left\|\frac{\Delta{\bm{w}}_j(t_k)}{\Delta t_k}\right\|,$$
where $\Delta t_k=t_k-t_{k-1}$, and $\Delta{\bm{w}}_j(t_k)={\bm{w}}_j(t_k)-{\bm{w}}_j(t_{k-1})$.

Apply Lemma \ref{Lemma2} and take the expectation w.r.t. ${D}_{t}$, and denote ${\bm{w}}_j(t_k)'=\frac{\Delta{\bm{w}}_j(t_k)}{\Delta t_k}$. Combined with the standard Rademacher complexity bound, we obtain the following error bound
\begin{equation}\label{A.D3}
	\begin{aligned}
		\frac{1}{K}\sum\limits_{t=1}^K\left(R(\mathbf{W}_t;\mathcal{D}_{t})-\hat R_l(\mathbf{W}_t)\right)\leq\frac{1}{K}2\sqrt 2 L\sum\limits_{j = 1}^C\sum\limits_{k=1}^K \Delta t_k\mathop\mathbb{E}\limits_{{D}_{t}\sim\mathcal{D}_{t}}\left\|{\bm{w}}_j(t_k)'\right\|+ B\sqrt{\frac{{\log (1/\delta )}}{2l}},
	\end{aligned}
\end{equation}	
Subsequently, we relate the Fisher-Rao distance \ref{PT3-1} and the generalization bound. For instance, if $\|\frac{\text{d}{\bm{w}}_j(t)}{\text{d}t}|$ is Riemann integrable over $t$
, then as $K$ goes to infinity, there exists a sequence $0=t_0\leq t_1\leq t_2,\cdots,\leq t_K\leq1$ such that 
$$\sum_{k=1}^K\Delta t_k \left\|\frac{\Delta{\bm{w}}_j(t_k)}{\Delta t_k}\right\|\to\int_0^1 {\left\|\dot{\bm{w}}_j(t)\right\|}\enspace\text{d}t=\int_0^1{\sqrt {\left\langle\dot{\bm{w}}_j(t),\bm{g}(\bm{w}_j(t))\dot{\bm{w}}_j(t)\right\rangle}}\enspace\text{d}t.$$
Theorem \ref{Theorem3} is proved.

\subsection{Discussion}

Looking back at the DAL framework in Eq. (\ref{DAL}), we have added the \textbf{Transport Model Reuse} and \textbf{Manifold Regularization} into the classification models, it is natural to ask the question: do these kinds of constraints improve performance? We will show that, together with the addition of these regularizers, the generalization ability of our algorithm will be improved.

The effect of the regularization term on the generalization ability of the DAL model is mainly on the hypothesis space. 
Subsequently, we discuss the different roles that \textbf{Model Reuse} and \textbf{Manifold Regularization} play in reducing the Rademacher complexity $\hat {\mathfrak{R}}_{l}(\bm{\mathcal H})$, respectively.

\subsubsection{Model Reuse Improves the Bound}

The model reuse regularization parameter $\alpha$ controls the amount that the pre-trained model constrains the hypothesis space. It's obvious from the definition of the hypothesis class $\bm{\mathcal H}$ that if $\alpha_1\geqslant\alpha_2\geqslant0$, then $\bm{\mathcal H}_{\alpha_1}\subseteq\bm{\mathcal H}_{\alpha_2}$, and thus $\hat {\mathfrak{R}}_{l}(\bm{\mathcal H}_{\alpha_1}) \leqslant \hat {\mathfrak{R}}_{l}(\bm{\mathcal H}_{\alpha_2})$. That is, increasing $\alpha$ reduces
the Rademacher complexity $\hat {\mathfrak{R}}_{l}(\bm{\mathcal H})$. 

Without loss of generality, let $\beta=0$, which makes it more intuitive to analyze the impact of model reuse on generalization ability. Then we obtain the reduced optimization problem
$$Q_1(\mathbf{W})=\frac{1}{l}\sum\limits_{i = 1}^{l} {\ell\left( {f\left( {\bm{x}_i} \right),{y}_i} \right)} + \alpha\left\| {{\mathbf{W}_t}-\bar {\mathbf{W}}_{t-1}} \right\|_F^2.$$
The reduced optimization problem $Q_1(\mathbf{W})$ reuses previous model $\bar {\mathbf{W}}_{t-1}$ as a biased regularizer, which ensures the current model ${\mathbf{W}_t}$ will not deviate far away from the provided $\bar {\mathbf{W}}_{t-1}$. We prove that the consideration of a well-trained model from a related homogeneous task facilitates the learning efficiency
in the current multi-class task. We collect this property in a proposition.
\begin{Proposition} \label{Proposition1}
	Considering the C-classification task with a linear model, let $\bm{\mathcal{W}_t}$ be the family of hypothesis set, and denote the hypothesis returned by the model trained with the current data and consistency constraint $\left\| {{\mathbf{W}_t}-\bar {\mathbf{W}}_{t-1}} \right\|_F^2$ as $\mathbf{W}_t$. 
	Suppose the loss function $\ell$ is $M$-bounded and $L$-Lipschitz w.r.t. Euclidean norm. Then, for any $\delta>0$, with probability at least $1-\delta$ over a labeled sample $S$ of size $l$, the following inequalities holds for all $\mathbf{W}_t\in \bm{\mathcal{W}_t}$
	\begin{equation}\label{tm1-1}
		R(\mathbf{W}_t;\mathcal{D}_{t}) \leqslant \hat R_{l}(\mathbf{W}_t)+\frac{2\sqrt {2}LC}{{\sqrt l }}\Lambda \sqrt{\frac{R(\bar {\mathbf{W}}_{t-1};\mathcal{D}_{t})}{\alpha}} + M \sqrt{\frac{{\log (1/\delta )}}{2l}}.
	\end{equation}
\end{Proposition}
Here, $R(\mathbf{W}_t;\mathcal{D}_{t})$ and $\hat{R}_{l}(\mathbf{W}_t)$ are the generalization error and empirical error. $\Lambda = \max\left\{ { \| \bm{x}  \|| \bm{x} \in \mathcal{X}} \right\}$ represents the radius of the feature domain. \\
%Besides, one more point should be mentioned here. In the above two propositions, we all need to assume that the loss function $\ell$ is $M$-bounded and $L$-Lipschitz w.r.t. Euclidean norm. Commonly, the $\textrm{CEL}(\cdot,\cdot)$ is in fact not $L$-Lipschitz and bounded \cite{aaai/WangG19}. To utilize this effective loss in the multi-class case, we use the same strategy as in \cite{2007Comprehensive} to make it satisfy these requirements. Concretely, in computing the prediction in Eq. (\ref{eq7}), we adopt the softmax (SF) output for the predictions. Then, $\textrm{CEL} \circ \textrm{SF} (\cdot, \cdot)$ is $(\sqrt{C+1}+1)$-Lipschitz.
\textbf{Proof of Proposition \ref{Proposition1}}~~~
	The proof of Proposition \ref{Proposition1} will start from the standard Rademacher complexity bound for $\bm{\mathcal{W}}_t$. Let $\bm{\mathcal{L}}$ be the family of loss function associated to $\bm{\mathcal{W}_t}$. For any $\delta > 0$, with probability at least $1-\delta$ over a sample $S$ of size $l$, the following inequality
	holds for all $\mathbf{W}_t\in \bm{\mathcal{W}_t}$:
	\begin{equation}\label{pro1}
		R_{\mathcal{D}_t}({\mathbf{W}_t})\leq\hat R_{l}({\mathbf{W}_t})+2{\mathfrak{R} _{l}}(\bm{\mathcal{L}})+M\sqrt{\frac{{\log (1/\delta )}}{2n_2}}
	\end{equation}
	
	Instead of learning the linear predictor directly, we focus on the helpfulness of reusing a related model $\bar {\mathbf{W}}_{t-1}$. So we can first bound the domain of the linear classifier $\mathbf{W}_t$. Denote the empirical optimal solution of the model as $\mathbf{W}^*_t$. Due to the optimality of the empirical objective at $\mathbf{W}_t$, we have	
	\begin{equation}\label{A4}
		\begin{aligned}
			\hat{R}_l\left(\mathbf{W}^*_t\right) +\alpha\left\|\mathbf{W}^*_t-\bar {\mathbf{W}}_{t-1}\right\|_F^2 \leq \hat{R}_l\left(\bar {\mathbf{W}}_{t-1}\right)+\alpha\left\|\bar {\mathbf{W}}_{t-1}-\bar {\mathbf{W}}_{t-1}\right\|_F^2=\hat{R}_l\left(\bar {\mathbf{W}}_{t-1}\right)
		\end{aligned}
	\end{equation}
	Thus $\left\|\mathbf{W}^*_t-\bar {\mathbf{W}}_{t-1}\right\|_F\leq\sqrt{\frac{\hat{R}_l\left(\bar {\mathbf{W}}_{t-1}\right)}{\alpha}}$. From Eq. (\ref{A4}), there is a loss constraint for the optimal solution $\mathbf{W}^*_t$, i.e., $\hat{R}_l\left(\mathbf{W}^*_t\right)\leq\hat{R}_l\left(\bar {\mathbf{W}}_{t-1}\right)$. In addition, we also have
	\begin{equation}
		\begin{aligned}
			\left\|\mathbf{W}^*_t\right\|_F =\left\|\mathbf{W}^*_t-\bar {\mathbf{W}}_{t-1}+\bar {\mathbf{W}}_{t-1}\right\|_F  \leq\left\|\mathbf{W}^*_t-\bar {\mathbf{W}}_{t-1}\right\|_F+\left\|\bar {\mathbf{W}}_{t-1}\right\|_F \leq \sqrt{\frac{\hat{R}_l\left(\bar {\mathbf{W}}_{t-1}\right)}{\alpha}}+\left\|\bar {\mathbf{W}}_{t-1}\right\|_F
		\end{aligned}
	\end{equation}
	So the optimal solution is in a bounded domain $$\mathbf{W}^*_t\in \bm{\mathcal{W}}_t=\left\{\mathbf{W}_t, \left\|\mathbf{W}^*_t-\bar {\mathbf{W}}_{t-1}\right\|_F\leq\sqrt{\frac{\hat{R}_l\left(\bar {\mathbf{W}}_{t-1}\right)}{\alpha}}, \hat{R}_l\left(\mathbf{W}^*_t\right)\leq\hat{R}_l\left(\bar {\mathbf{W}}_{t-1}\right)\right\}.$$
	
	Subsequently, we further analyze the second item of Eq.(\ref{pro1}). By the definition of Rademacher Complexity, we have
	\begin{equation}\label{A8}
		\begin{aligned}
			{\mathfrak{R} _{l}}(\bm{\mathcal{L}})&=\mathop{\mathbb{E}}\limits_{{\bm{\mathrm {Z}}} \sim {{\mathcal{D}}^{l}}}\left[ {E_\epsilon }\left[ {\mathop {\sup }\limits_{{\mathbf{W}_t} \in \bm{\mathcal{W}}_t} \sum\limits_{i = 1}^{l} {{\sigma _i}\ell({\mathbf{W}_t}({\bm{x}_i}))} } \right] \right] \mathop\leq\limits^{\textcircled{\small{1}}}\frac{\sqrt{2} L}{l} \mathbb{E}\left[\sup _{\mathbf{W} \in \mathcal{W}} \sum_{i = 1}^{l} \sum_{c=1}^C \epsilon_{i c} \bm{w}_c(t)^{\top} \bm{x}_i\right] \\
			&\mathop=\limits^{\textcircled{\small{2}}}\frac{\sqrt{2} L}{M} \mathbb{E}\left[\sup _{\mathbf{W} \in \mathcal{W}} \sum_{c=1}^C\left\langle\bm{w}_c(t), \sum_{i = 1}^{l} \epsilon_{i c} \bm{x}_i\right\rangle\right] \mathop=\limits^{\textcircled{\small{3}}}\frac{\sqrt{2} L}{M} \mathbb{E}\left[\sup _{\mathbf{W} \in \mathcal{W}} \sum_{c=1}^C\left\langle\bm{w}_c(t)-\bar{\bm{w}}_{c}(t-1), \sum_{i = 1}^{l} \epsilon_{i c} \bm{x}_i\right\rangle\right] \\
			& \leq \frac{\sqrt{2} L}{M} \mathbb{E}\left[\sup _{\mathbf{W} \in \mathcal{W}}\left\|\mathbf{W}_t-\bar {\mathbf{W}}_{t-1}\right\|_F \sum_{c=1}^C\left\|\sum_{i =1}^{l} \epsilon_{i c} \bm{x}_i\right\|\right] \leq \frac{\sqrt{2} L C}{M} \mathbb{E}\left[\sqrt{\frac{\hat R_l(\bar {\mathbf{W}}_{t-1})}{\alpha}}\right] \sqrt{l} \Lambda\\
			&\leq \frac{\sqrt{2} L C \sqrt{l} \Lambda}{M} \sqrt{\frac{R(\bar {\mathbf{W}}_{t-1};\mathcal{D}_{t})}{\alpha}}
		\end{aligned}
	\end{equation}
	In above derivations, ${\textcircled{\small{1}}}$ comes from the fact that the loss function $\ell$ is $L$-Lipschitz. ${\textcircled{\small{2}}}$ results from the linearity of inner product. In ${\textcircled{\small{3}}}$, we introduce the constant vector $\bar{\bm{w}}_{c}(t-1)$, which is the $c$-th column of prior weight matrix $\bar {\mathbf{W}}_{t-1}$. It is notable that since this is a constant term, when combined with Rademacher variables it only outputs zero.	
	
Proposition \ref{Proposition1} provides a generalization error bound closely related to the expected risk of previous models. If the provided model $\bar {\mathbf{W}}_{t-1}$ adapts well on the current task distribution, i.e., a small expected risk $R(\bar{\mathbf{W}}_{t-1};\mathcal{D}_{t})$, such that the r.h.s. of Eq.( \ref{tm1-1}) will be more compact. Here $R(\bar {\mathbf{W}}_{t-1};\mathcal{D}_{t})$ naturally acts as a task relatedness measure. Thus, reusing a suitable related model helps reduce the hypothesis space complexity for the target learning problem, and facilitates the current task learning efficiency. In other words, with limited current task labeled examples, the current learned model $\mathbf{W}_t$ can achieve higher performance in expectation.
\subsubsection{Manifold Regularization Improves the Bound}
Similarly, the manifold regularization parameter $\beta$ controls the amount that the manifold regularization constrains the hypothesis space. It's obvious from the definition of the hypothesis class $\bm{\mathcal H}$ that if $\beta_1>\beta_2>0$, then $\bm{\mathcal H}_{\beta_1}\subseteq\bm{\mathcal H}_{\beta_2}$, and thus $\hat {\mathfrak{R}}_{l}(\bm{\mathcal H}_{\beta_1}) \leqslant \hat {\mathfrak{R}}_{l}(\bm{\mathcal H}_{\beta_2})$. That is, increasing $\beta$ reduces
the Rademacher complexity $\hat {\mathfrak{R}}_{l}(\bm{\mathcal H})$. The amount of
this reduction is characterized by the expression
$$\Delta (\beta ): = \tilde{\beta}\textrm{Tr}\left(\mathbf{J}^\top(\mathbf{I}+\tilde{\beta}\mathbf{M})\mathbf{J}\right)$$ 
from Theorem \ref{Theorem2}, where $\tilde{\beta}=\frac{C\beta}{\hat R_l(\bar{\mathbf{W}}_{t-1})}$. When $\beta=0$, the algorithm ignores the unlabeled data, and the reduction in indeed $\Delta(0)=0$. As we would expect, $\Delta(\beta)$ is nondecreasing in $\beta$ and has a finite limit as $\beta\to \infty$. We collect these properties in a proposition:
\begin{Proposition} \label{Proposition2}
	Suppose that the inverse of $\mathbf{M}$ exists. $\Delta(0)=0$, $\Delta(\beta)$ is nondecreasing on $\beta\geq0$, and 
	$$\mathop {\lim }\limits_{\beta  \to \infty } \Delta(\beta)=\textrm{Tr}\left(\mathbf{J}^\top\mathbf{M}^{-1}\mathbf{J}\right).$$
\end{Proposition}
Here $\mathbf{J}=\tilde{\alpha}^{-1}\mathbf{G}^\top\mathbf{K}_l^\top,\enspace \mathbf{M}=\tilde{\alpha}^{-1}\mathbf{G}^\top\mathbf{K}\mathbf{G}$, which is consistent with Theorem \ref{Theorem2}.\\
\textbf{Proof of Proposition \ref{Proposition2}}~~~
The limit claim is clear if we write the reduction as $$\Delta(\beta)=\textrm{Tr}\left(\mathbf{J}^\top(\tilde{\beta}^{-1}\mathbf{I}+\mathbf{M})^{-1}\mathbf{J}\right).$$
	Since $\mathbf{K}$ is Gram matrix, $\mathbf{M}$ is positive semidefinite (psd). Thus we can take an orthogonal decomposition of $\mathbf{M}$, i.e., $\mathbf{M}=\mathbf{P}^\top\mathbf{\Sigma}\mathbf{P}$,with diagonal $\mathbf{\Sigma}>0$ and orthogonal $\mathbf{P}$. Then
	\begin{equation}
		\Delta(\beta)=\textrm{Tr}\left(\mathbf{J}^\top\mathbf{P}^\top(\tilde{\beta}^{-1}\mathbf{I}+\mathbf{\Sigma})^{-1}\mathbf{P}\mathbf{J}\right)=\sum\limits_{i = 1}^N\sum\limits_{j = 1}^N {\left(PJ\right)}_{ij}^2\left( {\tilde{\beta}^{-1}+\mathbf{\Sigma}_{jj}} \right)^{-1}.
	\end{equation}
	From this expression, it's clear that $\Delta(\beta)$ is nondecreasing in $\beta$ on $(0,\infty)$. Since $\Delta(\beta)$ is continuous at $\beta = 0$, it's nondecreasing on $\left[ {0,\infty} \right)$.

Taking together Theorem \ref{Theorem2} and Proposition \ref{Proposition2}, we can draw a conclusion that is consistent with the intuition that manifold regularization helps reduce the hypothesis space complexity and facilitates the current task learning efficiency, while the reduction has a finite limit as $\beta \to \infty$. 

Loosely summarized, the model reuse and semi-supervised regularization improve the generalization bound by reducing the Rademacher complexity $\hat {\mathfrak{R}}_{l}(\bm{\mathcal H})$. The trade-off parameters $\alpha$ and $\beta$ control the amount that the pre-trained models and the amount that the manifold regularization constrains the hypothesis space, respectively. The integration of these two components facilitates the performance enhancement of DAL.

\section{Framework Implementations}

\subsection{Two Illustrations}

The above model in Eq. (\ref{DAL framework}) only provides a general framework. In real applications, we should consider concrete forms for implementation. There are many surrogate loss functions such as least square loss, hinge loss, cross-entropy loss, etc. Considering that Cross Entropy Loss (CEL) \cite{de2005tutorial} has been deemed as a representative loss for multi-class classification, and the Least Square (LS) loss has the convenience for optimization solution. We chose CEL and LS for illustration. 

First, we show the empirical loss under these two losses. For brevity, we omit the index $t$ of the data matrix $\mathbf{X}_t$ at each evolving moment.
Denote the $j$-th column (corresponding to the $j$-th class) of $\mathbf{W}_t$ as ${\bm{w}}_j(t)$. The predicted label for the $j$-th class of $\mathbf{x}_i$ is defined as $\hat{y}_{ij}$ for $i=1, 2, \cdots, l$. If we use the softmax strategy, it has the following form.
\begin{equation} \label{softmax}
	\begin{split}
		\hat{y}_{ij} = \frac{\exp(({\bm{w}}_j(t))^\top\bm{x}_i)}
		{\sum_{j=1}^{C} \exp(({\bm{w}}_j(t))^\top\bm{x}_i)}.
	\end{split}
\end{equation}
The CEL loss function becomes
\begin{equation} \label{eq8}
	\begin{split}
		-\sum_{i=1}^{L} \sum_{j=1}^{C} {y}_{ij} \log \left( \hat{y}_{ij} \right).
	\end{split}
\end{equation}
Here, ${y}_{ij} \in \{0, 1\}$ is the true label for the $j$-th class of the $i$-th training point in the C-stage of each moment during the evolution. 

Consider the same linear classifier $\mathbf{W}_t$, the LS loss is formulated as:
$$\left\|\mathbf{W}_t^\top\mathbf{X}+\textbf{b}\bm{1}^\top-\mathbf{Y}\right\|_F^2,$$
where $\textbf{b}\in \mathbb{R}^C$. Using the fact that $\textbf{b} = \frac{1}{l}(\mathbf{Y}\bm{1} - \mathbf{W}_t^\top\mathbf{X}\bm{1})$, we can introduce the centralization matrix $\mathbf{H} = \mathbf{I} - \frac{1}{l}\bm{1}\bm{1}^\top $, and get rid of the bias vector $\textbf{b}$ as $\left\|\mathbf{W}_t^\top\mathbf{X}\mathbf{H}-\mathbf{Y}\mathbf{H}\right\|_F^2.$

As for manifold regularization, we adopt the most commonly used graph regularization for illustration. Combining the \textbf{Transport Model Reuse} and the \textbf{Manifold Regularization}, we consider the following two optimization problems for our DAL approach.

\emph{A. DAL-LS}
\begin{equation}\label{DAL-LS}
	\begin{aligned}
		\mathop {\min }\limits_{\mathbf{W}_t \in \mathcal{W}_t} \left\|\mathbf{W}_t^\top\mathbf{X}\mathbf{H}-\mathbf{Y}\mathbf{H}\right\|_F^2 + \alpha\left\| {{\mathbf{W}_t}-d\mathbf{T}\mathbf{W}_{t-1}} \right\|_F^2
		+\beta \textrm{Tr}({\mathbf{W}_t}^\top\mathbf{X})^\top\mathbf{G}({\mathbf{W}_t}^\top\mathbf{X}).
	\end{aligned}
\end{equation}
where $\alpha>0$ and $\beta>0$ are two parameters. For this illustration, we employ the LS loss to train the model. Hence, we name this DAL method DAL-LS. 

\emph{B. DAL-CEL}
\begin{equation}\label{DAL-CEL}
	\begin{aligned}
		\mathop {\min }\limits_{\mathbf{W}_t \in \mathcal{W}_t} -\sum_{i=1}^{L} \sum_{j=1}^{C} {y}_{ij} \log \left( \hat{y}_{ij} \right)  + \alpha\left\| {{\mathbf{W}_t}-d\mathbf{T}\mathbf{W}_{t-1}} \right\|_F^2+\beta \textrm{Tr}({\mathbf{W}_t}^\top\mathbf{X})^\top\mathbf{G}({\mathbf{W}_t}^\top\mathbf{X}).
	\end{aligned},
\end{equation}
For this illustration, we employ the representative CEL loss to train the model. Hence, we name this DAL method DAL-CEL. 

Comparing the above two implementations, we have the following mentioned points. (1) The formulations in Eqs. (\ref{DAL-LS}) and (\ref{DAL-CEL}) adopt two classical losses to train the model, respectively. They will take different effects on model training, from both the theoretical and experimental aspects. (2) When reusing the evolving distribution model, DAL-LS and DAL-CEL adopt KME or KDE to encode marginal distribution information of features for computing the evolving cost matrix.

\subsection{Optimization}

There are two detailed illustrations in Eqs. (\ref{DAL-LS}) and (\ref{DAL-CEL}). For MTS-LS, it is easy to obtain a closed-form solution due to the convenience of LS loss for optimization. Specifically, since each term in Eq. (\ref{DAL-LS}) is convex w.r.t. $\mathbf{W}_t$, there is a global optimal closed-form solution for $\mathbf{W}_t$.
$$\mathbf{W}_t = \left(\mathbf{X}\mathbf{H}\mathbf{X}^\top+\alpha\mathbf{I}+\beta\mathbf{X}\mathbf{G}\mathbf{X}^\top\right)^{-1}\left(\mathbf{X}\mathbf{H}\mathbf{Y}^\top+\alpha d \mathbf{T}\mathbf{W}_{t-1}\right).$$
For DAL-CEL, since we have utilized CEL to train the model, it is difficult to derive a closed-form solution. An iterative algorithm is designed for optimization. Regarding the optimization of DAL-LS, we will refrain from reiterating the specifics and instead focus on a comprehensive analysis of the solution for DAL-CEL.

\subsubsection{Optimization Algorithm for DAL-CEL}

We use the gradient descent algorithm to solve this problem since the CEL loss can be tackled in an alternative way. For the sake of presentation and without ambiguity, we denote the problem in Eq. (\ref{DAL-CEL}) as
\begin{equation} \label{optimization1}
	\normalsize
	\min_{\mathbf{W}_t}~J(\mathbf{W}_t)= \mathop {\min }\limits_{\mathbf{W}_t \in \mathcal{W}_t} -\sum_{i=1}^{L} \sum_{j=1}^{C} {y}_{ij} \log \left( \hat{y}_{ij} \right)+ \alpha\left\| {{\mathbf{W}_t}-d\mathbf{T}_1\mathbf{W}_{t-1}} \right\|_F^2
	+\beta \textrm{Tr}({\mathbf{W}_t}^\top\mathbf{X})^\top\mathbf{G}({\mathbf{W}_t}^\top\mathbf{X}).
\end{equation}
where $\mathbf{W}_t \in \mathbb{R}^{d \times C}$. $y_{ij}$ is the true label for the $j$-th class of ${\bm {x}}_i \in \mathbb{R}^{d}$. $\hat y_{ij}$ is the predicted label for the $j$-th class of ${\bm {x}}_i$, whose definition is shown in Eq. (\ref{softmax}).

Denote $\mathbf{Z} = [Z_{ij}] = \mathbf {X}_l \mathbf{W}_t \in \mathbb{R}^{l \times C} $ and $J_1(\mathbf{W}_t)$ as follows,
\begin{equation}
	J_1(\mathbf{W}_t)=-\sum_{i=1}^{l}\sum\limits_{j = 1}^C {y}_{ij} \log \left( \hat{y}_{ij} \right)
\end{equation}
The key step for the gradient descent algorithm is to calculate the gradient with respect to $\mathbf{W}_t$.

Without loss of generality, we set $\log \left( \hat{y}_{ij} \right) =\ln \left( \hat{y}_{ij} \right)$. Then, we have $Z_{ij} = \sum_{k=1}^{d}X_{ik}W_{kj}$. For any $w_{pq}~ (1\leq p\leq d, 1\leq q\leq C)$, we can obtain
\begin{equation}
	\frac{\partial J_1(\mathbf{W}_t)}{\partial w_{pq}} = \sum_{i=1}^{l} \frac{\partial J_1}{\partial Z_{iq}}\frac{\partial Z_{iq}}{\partial w_{pq}}
\end{equation}
\begin{equation}\label{optimization2}
	\frac{\partial J_1(\mathbf{W}_t)}{\partial Z_{iq}}=\frac{\partial \sum_{u=1}^{l}(-\sum_{j=1}^{C}{y}_{ij} \ln\left( \hat{y}_{ij} \right))}{\partial Z_{iq}}= -\sum_{j=1}^{C}\frac{{y}_{ij}}{\hat{y}_{ij}}\frac{\partial\hat{y}_{ij}}{\partial Z_{iq}}.
\end{equation}
Here, the second equality holds since the gradient is nonzero if and only if $u=i$.  Recall the definition of ${\hat{y}_{ij}}$, we have 
\begin{equation} \label{optimization3}
	\frac{\partial\hat{y}_{ij}}{\partial Z_{iq}}=\left\{
	\begin{aligned}
		&\frac{e^{Z_{ij}}}{\sum_{c=1}^{C+1}e^{Z_{ic}}}-\frac{e^{Z_{ij}}e^{Z_{iq}}}{(\sum_{c=1}^{C+1}e^{Z_{ic}})^2}=\hat{y}_{iq}-(\hat{y}_{iq})^2 &j=q \\
		&-\frac{e^{Z_{ij}}e^{Z_{iq}}}{(\sum_{c=1}^{C+1}e^{Z_{ic}})^2}=-\hat{y}_{ij}\hat{y}_{iq}                 &j\neq q \\
	\end{aligned}
	\right.
\end{equation}

Plugging Eq. (\ref{optimization3}) into Eq. (\ref{optimization2}), since $\sum_{j=1}^{C}y_{ij}=1$, we have
\begin{equation} \label{optimization4}
	\frac{\partial J_1(\mathbf{W}_t)}{\partial Z_{iq}}=\hat{y}_{iq}-y_{iq}.
\end{equation}

Note that $Z_{iq} = \sum_{k=1}^{d}X_{ik}W_{kq}$, plugging Eq. (\ref{optimization4}) into Eq. (\ref{optimization2}), we have
\begin{equation}\label{optimization5}
	\begin{split}
		\frac{\partial J_1(\mathbf{W}_t)}{\partial W_{pq}}=\sum_{i=1}^{l}\frac{\partial J_1(\mathbf{W}_t)}{\partial Z_{iq}}X_{ip}=\sum_{i=1}^{l}(\hat{y}_{iq}-y_{iq})X_{ip}.
	\end{split}
\end{equation}

In summary, we can have the following derivation

\begin{equation} \label{derivation1}
	\begin{split}
		\frac{\partial J(\mathbf{W}_t)}{\partial {\mathbf{W}_t}} =\nabla J_1 + 2\alpha (\mathbf{W}_t-d\mathbf{T}_1\mathbf{W}_{t-1})+2\beta\mathbf{X}^\top\mathbf{G}\mathbf{X}\mathbf{W}_t,
	\end{split}
\end{equation}

Then, we can have the following updating rule.
\begin{equation}  \label{optimization6}
	\begin{split}
		\mathbf{W}_{t+1}=\mathbf{W}_{t}-\eta\nabla J(\mathbf{W}_{t}).
	\end{split}
\end{equation}
where $\eta>0$ is the updating step size. 

In summary, the procedure in optimizing Eq. (\ref{DAL-CEL}) is summarized in Algorithm \ref{alg1}.
\begin{algorithm}[]
	\caption{Algorithm in Solving DAL Problems in Eq. (\ref{DAL-CEL}).}
	\label{alg1}
	\begin{algorithmic}
		\STATE Input: $\mathbf{X}$, $\mathbf{Y}$, $\alpha$, $\beta$, $\eta$, $\lambda$.
		\STATE \textbf{Repeat}
		\STATE 1: Calculate the the evolving cost matrix $\mathbf{C}$ by Eq. (\ref{EC}) or Eq. (\ref{EC1}). Then, learn a transport matrix  $\mathbf{T}$ by solving the optimization problem in Eq. (\ref{OT}).
		\STATE 2: Calculate Laplacian matrix by $\bm{\mathrm {G}}=\bm{\mathrm {D}} - \bm{\mathrm{A}}$, where $\bm{\mathrm{A}}$ is calculated by Eq. (\ref{LP}), and $\bm{\mathrm {D}}$ is a diagonal matrix with diagonal elements ${{D}_{ii}}=\sum\nolimits_{j = 1}^n {{{A}_{ij}}}$. 
		\STATE 3:  Initialize $\mathbf{W}_t$
		\STATE 4: Update $\mathbf{W}_t$ using the rules defined in Eq. (\ref{optimization6}), where $ \nabla J({\mathbf{W}}_t)$ is defined in Eq. (\ref{derivation1}).
		\STATE \textbf{Until converges}\\
		\STATE Output: The optimal classifier $\mathbf{W}_t$ for the new coming data.\\
		\textbf{End procedure}
	\end{algorithmic}
\end{algorithm}

\subsubsection{Convergence Analysis}

We solve the problem in Eq. (\ref{DAL-CEL}) alternatively by the strategy in Algorithm \ref{alg1}. The following theoretical results guarantee the effectiveness of our implementations.
\begin{Proposition} \label{Proposition3}
	The Algorithm \ref{alg1} can find the global optimal solution to the problem in Eq. (\ref{DAL-CEL}) when the step size $\eta$ is small enough.
\end{Proposition}
The sketch of proof is presented in a standard manner. Given that the procedure in Algorithm \ref{alg1} follows a gradient descent style, it guarantees global optimization when the objective function is convex and the step size is sufficiently small \cite{BV2014}.

In Eq. (\ref{DAL-CEL}), or its counterpart in Eq. (\ref{optimization1}), we need to check that the objective function $J(\mathbf{W}_t)$ is convex w.r.t. $\mathbf{W}_t$. Obviously, the CEL has been proven to be convex \cite{BoerKMR05}. The second term, i.e., $\left\| {{\mathbf{W}_t}-d\mathbf{T}_1\mathbf{W}_{t-1}} \right\|_F^2$ is also convex. The last term, i.e., $\textrm{Tr}({\mathbf{W}_t}^\top\mathbf{X})^\top\mathbf{G}({\mathbf{W}_t}^\top\mathbf{X})$. Its Hessian matrix w.r.t. $\mathbf{W}_t$ is semi-definite since $\mathbf{G}$ is semi-definite and it is also convex. Finally, the objective function in Eq. (\ref{optimization1}) is their weighted sum with a positive weighting coefficient. Thus, it is convex.

The following points deserve emphasis: (1) While the aforementioned results provide theoretical guarantees on convergence behavior, it is essential to ensure that the step size $\eta$ is sufficiently small. In practical implementations, we employ a useful strategy as follows. Initially, we select a slightly larger step size (0.1 in our experiments). Subsequently, if there is only a small reduction in the objective function value, we decrease the step size by multiplying it with a shrinking factor (0.5 in our experiments). (2) The experimental results demonstrate the rapid convergence of our algorithms. Consequently, we terminate the iteration process if either the number of iterations exceeds 1000 or the change in objective function value falls below $10^{-6}$.

\section{Related Works}

The first related learning paradigm is domain adaptation learning. In the literature, there are several pioneer works. These methods can be roughly divided into two categories. The first type of method is based on distribution discrepancy \cite{DBLP:journals/tnn/PanTKY11,DBLP:conf/iccv/LongWDSY13,DBLP:conf/icdm/WangCHFS17,DBLP:conf/ijcnn/ZhangW20}, which directly minimizes the distance of source and target domain distributions through transformation. For example, transfer component analysis (TCA) \cite{DBLP:journals/tnn/PanTKY11} adopted maximum mean discrepancy
(MMD) \cite{DBLP:journals/jmlr/GrettonBRSS12} to reduce the discrepancy across domains and preserve data information. The second type of method is based on subspace learning \cite{DBLP:journals/ijcv/ShaoKF14,DBLP:journals/tip/XuFWLZ16,DBLP:journals/kbs/RazzaghiRA19}, which assumes that different domains lie in the common transformed subspace. For instance, Shao et al. \cite{DBLP:journals/ijcv/ShaoKF14} proposed low-rank transfer subspace learning (LTSL), which pointed out that the reconstruction matrix should has blockwise structure. Although these approaches have achieved prominent performances in their application fields, they differ from DAL in the following aspects. (1) The motivations are different. All these approaches assume that there are two discrete domains, without considering the evolving target domains. Mainstream domain adaptation methods are tailored to adapting discrete source and target domains \cite{DBLP:journals/ml/Ben-DavidBCKPV10,DBLP:conf/nips/Ben-DavidBCP06,DBLP:journals/corr/abs-2010-03978,DBLP:conf/ijcai/0001LLOQ21}. The evolving data distribution (domains) poses obstruction to knowledge transfer. Thus, These methods are not suitable for our investigations. (2) The traditional approaches focus on the model design for adapting to the target domain, while our approach also deepens the understanding of evolving data distribution from a theoretical aspect. The theoretical results guide the design of the DAL model framework to address this crucial but rarely studied problem.

Another related paradigm is model reuse, which is the key factor for Learnware \cite{DBLP:journals/fcsc/Zhou16a}. For example, Ye et al. have proposed the REctiFy via heterOgeneous pRedictor Mapping (REFORM) framework, which offers a solution for reusing models across different features or labels \cite{pami/YeZJZ21}. Yang et al., have proposed a comprehensive model reuse scheme called Fixed Model Reuse (FMR), which effectively harnesses the learning capabilities of deep models to implicitly extract valuable discriminative information from commonly used fixed models/features in general tasks \cite{aaai/YangZFJZ17}. Wu et al., have proposed a method known as Heterogeneous Model Reuse (HMR) to optimize the multiparty multiclass margin, which is defined to measure the global behavior of heterogeneous local models. Through this way, they effectively leverage local models to approximate a global model \cite{icml/WuLZ19}. Zhang et al. postulated the availability of a plethora of models in the learning software market and proposed an approach to effectively leverage acquired knowledge even when it is confronted with novel components \cite{aaai/ZhangY0Z21}. These prominent approaches can not tackle our problem either since (1) The violation of independent and identically distributed assumption poses an obstruction to model reuse. Mainstream model reuse methods are based on the independent and identically distributed assumption, which is different from our scenario. Thus, they cannot be borrowed for our task directly. (2) Although some other model reusing strategies primarily address scenarios involving heterogeneous feature spaces, which is also not aligned with our scenario. It will render them inapplicable to the problem at hand.

\section{Experiment}\label{s6}

In this section, we will evaluate the performance of our DAL approach with other closely related methods in different aspects. There are totally four groups of experiments. In the first group, we compare DAL with the comparison methods on the toy example. In the second group, we present the performances on several benchmark datasets. In the third group, we report results concerning the deep analyses of DAL in different aspects, including the trend of model performance with the evolving tasks and convergence behavior. Finally, we apply our approach to the application of digit recognition. Before going into the details, let us introduce data sets and baselines first.
\renewcommand\arraystretch{0.8}
\renewcommand\tabcolsep{18pt} %列间距
\begin{table}[h]
	\centering
	\caption{A brief description of data sets.}
	\normalsize
	%\scriptsize
	\begin{tabular}{ccccc}
		\toprule[1pt]
		Dataset& Instance & Feature & Class & Task\\
		\midrule[ .75pt]
		Spam & 9324 & 500 & 2 & 4\\	
		Weather & 18159 & 8 & 2 & 4\\
		Electricity & 45312 & 8 & 2 & 4\\
		PIE & 41268 & 1024 & 68 & 4\\
		VLSC & 10729 & 4096 & 5 & 3\\
		AWDC & 2533 & 800 & 10 & 3\\
		Amazon Review & 7996 & 400 & 2 & 3\\
		Cross-dataset Testbed & 10472 & 4096 & 40 & 2\\
		Waveform & 5000 & 21 & 3 & 4\\
		Magic & 10000 & 10 & 2 & 4\\
		Poker hand & 25010 & 10 & 10 &4\\
		\bottomrule[1pt]
	\end{tabular}
	\label{Data Introduction}
	\vskip -.2in
\end{table}

\subsection{Configuration}\label{s6.1}

Note that when dealing with real-world datasets, we cannot grasp the evolving distribution, like the start and end time of drift, and the underlying distribution. Therefore, both toy and synthetic datasets are included in the comparison experiments. One toy example and eight synthetic data sets are used in the experiments, i.e., Spam, Weather, Electricity, PIE, VLSC, AWDC, Amazon Review, Cross-dataset Testbed, Waveform, Magic, Poker hand, which arise from different applications. \\
\textbf{Dataset Generation.}~~~As mentioned above, during the evolution process, the data come in batches sequentially $T=\{D_0, D_1,\cdots,D_T \}$. In the experiments, each batch corresponds to a classification task, so there is a flow of tasks under an evolving distribution. We adaptively transform the existing dataset. For each domain adaptation dataset, we treat each domain as a task and permutate multiple domains to obtain a distribution-evolving task flow, such as VLSC. For the rest dataset, the data are sorted according to their arrival time and their distribution gradually changes as time passes, so we partition the dataset according to the arrival time to obtain a task flow with evolving data distribution.
Specifically, each dataset is divided into multi tasks, where task 0 is used as the initiation stage to train a well-trained related classifier with twice the sample size of other tasks. From the instance level, apart from the initiation stage, which is fully labeled, subsequent tasks only possess 1\% of annotated data. The following is a detailed introduction of the synthetic data sets, together with a description in Table \ref{Data Introduction}. 

\textbf{Spam}\footnote{http://spamassassin.apache.org} consists of 9324 email messages extracted from the Spam Assassin Collection. Each email is classified as either spam or legitimate and represented by a boolean bag-of-words approach with 500 binary features. The emails are sorted according to their arrival time and their distribution gradually changes as time passes, so this dataset represents gradual distribution evolution. 

\textbf{Weather} includes 18159 daily readings
composed of features such as temperature,
pressure, and wind speed. We use the same 8
features as \cite{DBLP:journals/tnn/ElwellP11}. Each instance is labeled as either rain or no rain. This dataset exhibits recurrent and gradual distribution evolution.

\textbf{Electricity} \cite{DBLP:journals/tkde/ZhaoWXGZ21} is widely adopted and is collected from the Australian New South Wales Electricity Market where prices are affected by demand and supply of the market. The dataset contains 45,312 instances with 8 features. The class label identifies the change in the price relative to a moving average of the last 24 hours. The data are sorted according to their arrival time and their distribution gradually changes as time passes, so this dataset represents gradual distribution evolution. 

\textbf{PIE} \cite{DBLP:conf/iccv/LongWDSY13} is a benchmark for face recognition. The database has 68 individuals with 41,368 32×32 face images. It has five subsets: C05 (left pose), C07 (upward pose), C09 (downward pose), C27 (frontal pose), and C29 (right pose). In each subset (pose), all face images were taken under different lighting, illumination, and expression conditions. 

\textbf{VLSC} \footnote{https://github.com/jindongwang/transferlearning/tree/master/data} VOC2007, LabelMe, SUN09, and Caltech101 are the object recognition datasets. Specifically, VOC2007 dataset contains four major categories and 5,011 images in total. It can subdivide four categories into 20 subcategories: airplane, bicycle, bird, sofa, etc. LabelMe contains 455 categories and each with 3342 images at least. SUN09 dataset contains 12000 labeled pictures covering rich shooting scenes and involves more than 200 categories. Caltech101 has 8677 images from 101 classes. In order to realize knowledge transfer between these datasets, 5
common categories are selected: bird, car, chair,
dog and person. 

\textbf{AWDC}\footnote{https://github.com/jindongwang/transferlearning/tree/master/data} Office+Caltech-256 is a popular benchmark for visual DA. Office contains 31 categories and 4652 pictures of office supplies, such as backpack, mouse, monitor, etc. It contains three subdatasets: Amazon, Webcam, and DSLR. Each category of
Amazon has 90 images on average, while Webcam
and DSLR has 30 images on average for each category. Caltech-256 keeps 256 categories with 80 pictures at least and 30607 in total. Gong et al. \cite{DBLP:conf/cvpr/GongSSG12} select the same category in Office and Caltech-256 to form the Office+Caltech10 dataset, with 2533 samples and 10 categories. 

\textbf{Amazon Review (AR)} \footnote{https://github.com/jindongwang/transferlearning/tree/master/data} contains reviews of four types of products: books, DVDs, electronics, and kitchen appliances. Each domain has around 2000 samples belonging to 2 classes: positive or negative \cite{DBLP:conf/icml/ChenXWS12}. 

\textbf{Cross-Dataset Testbed (C-DT)} \footnote{https://github.com/jindongwang/transferlearning/tree/master/data} is a Decaf7-based cross-dataset image classification dataset. It contains 40 categories of images from 3 domains: 3,847 images in Caltech256(C), 4,000 images in ImageNet(I), and 2,626 images for SUN(S). 

\textbf{Waveform, Magic, Poker hand } are all available at UCI Archive \footnote{http://archive.ics.uci.edu/datasets.html}.
Waveform consists of 5000 instances with 21 attributes. The instances include noise and 3 classes of waves have been generated from a combination of two of the three base waves. Magic consists of 10000 instances with 10 attributes. The data set was generated by a Monte Carlo program and divided into two categories. Poker hand consists of 25010 instances for 10 classes with 10 attributes. Each record is an example of a hand consisting of five playing cards drawn from a standard deck of 52. 
\\
\textbf{Baselines.}~~~As seen from the formulation of our DAL algorithm in Eq. (\ref{DAL-LS}) and Eq. (\ref{DAL-CEL}), we have two concrete formulations for our algorithm. For the sake of presentation, we name DAL with Cross Entropy Loss and Least Square loss as DAL-CEL and DAL-LS, respectively. Besides, since this is a new setting, there is rare work considering this task and we compare our DAL-CEL and DAL-LS with three groups of baseline methods with different purposes, i.e., TCA, JDA, W-BDA, DTLS, LS-OT, LS-G, and LSSVM \cite{DBLP:journals/jmlr/YeX07}. (1) To show the effectiveness of our DAL model in tracking evolving distribution, we compare DAL with the traditional domain adaptation methods, i.e., TCA \cite{DBLP:journals/tnn/PanTKY11}, JDA \cite{DBLP:conf/iccv/LongWDSY13}, W-BDA \cite{DBLP:conf/icdm/WangCHFS17}, DTSL \cite{DBLP:journals/tip/XuFWLZ16}, which are tailored to adapting discrete source and target domains. In the evolving data distribution scenario, the domain adaptation methods may suffer from a lack of annotated source domain data, as only the data in the initialization stage is fully annotated and there is an evolution of task flow. A straightforward strategy is employed to eliminate the obstruction. Specifically, the data from the former task, along with its predicted labels obtained after completing the previous classification task, are utilized as source domain data for subsequent domain adaptation. (2) To illustrate the effectiveness of the transport model reuse, we compare DAL with the traditional supervised learning methods, which train the model with limited current labeled data. This kind of approach does not consider the reusability of related models and the evolution of data distribution. The present study employs least square SVM (LSSVM) \cite{DBLP:journals/jmlr/YeX07} as an illustration. (3) To validate the effectiveness of semi-supervised regularization and transport model reuse, we compare DAL with LS-OT and LS-G for ablation experiments. The LS-OT and LS-G are specific cases of our approach, representing the implementation of the DAL framework with solely transport model reuse or exclusively semi-supervised regularization terms, respectively. 
\\
\textbf{Generating Well-Trained Source Model.}~~~A least square SVM (LSSVM) \cite{DBLP:journals/jmlr/YeX07} is trained on the initiation stage to get a well-trained Source Model, with parameter tuned by cross-validation.
\\
\textbf{Parameter Setting.}~~~The current task is limited in annotated data availability, thus default parameters are employed for all methods. For instance, in the case of DAL, the values of parameters $\alpha$ and $\beta$ are set to 1 and 0.1 respectively. As for the comparison methods, we utilize the suggested default parameter settings provided by their original authors. This parameter configuration is also applied to other experiments. Finally, in the following experiments, classification accuracy (Acc) is employed as the evaluation metric.
\begin{figure*}[!t]
	\setlength{\abovecaptionskip}{-.1cm}
	\setlength{\belowcaptionskip}{-.1cm}
	\begin{center}
		\includegraphics[width=13.5cm]{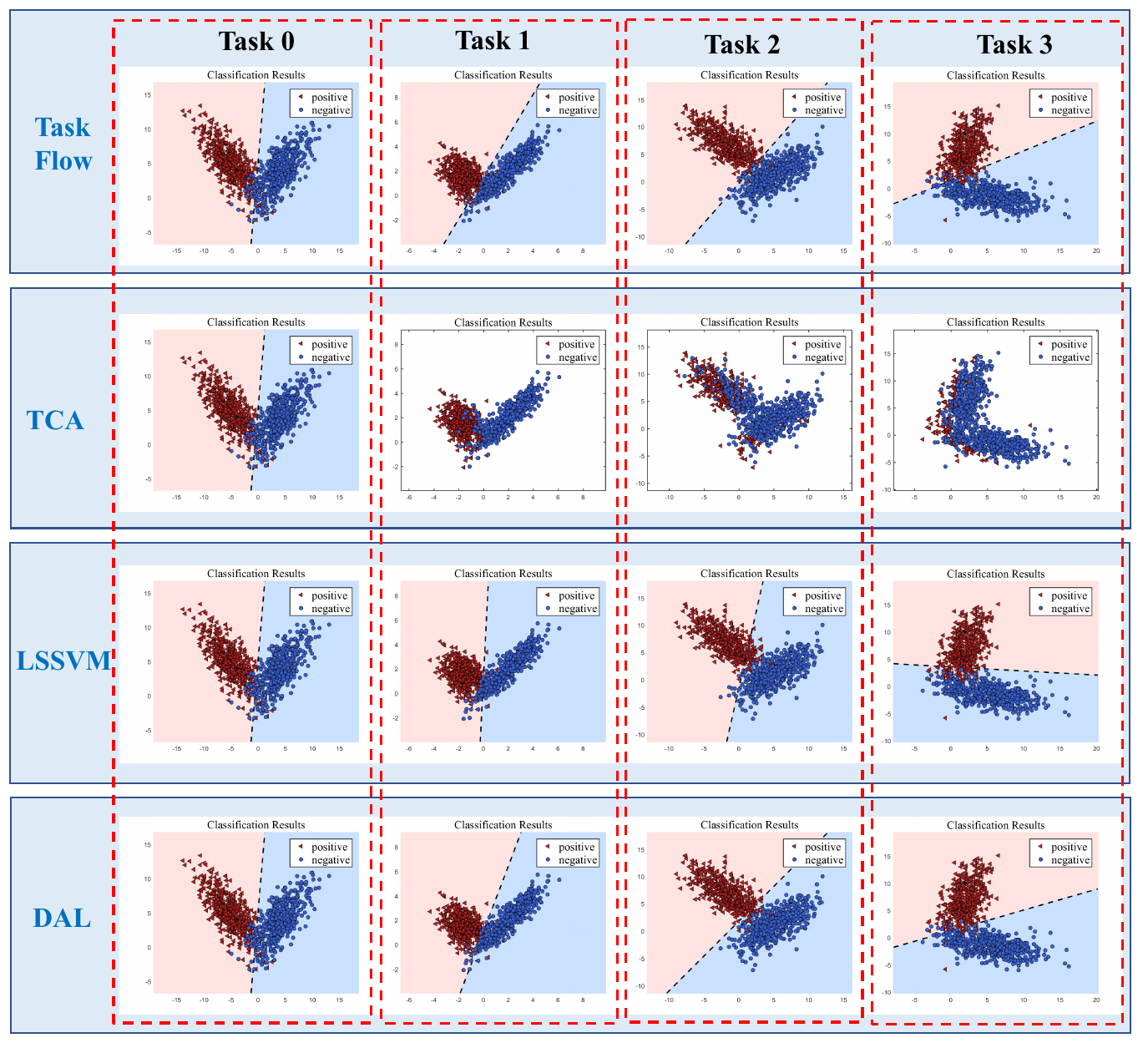}
	\end{center}
	\caption{Experimental results of DAL and representative comparison methods on a toy example. As depicted in the first row, encompasses four distinct classification tasks characterized by diverse data distributions. The following three rows correspond to the classification results of the two comparison methods and DAL. Task 0 represents the initiation stage and involves fully labeled data for training the source model. Note that Task 0 is the starting stage and all methods stand at the same starting line (Source model Result).}
	\label{Toy example}
\end{figure*}

\subsection{Toy Example}\label{s6.2}

To demonstrate the efficacy of the DAL model in tracking evolving data distributions more intuitively, we generate a two-dimensional toy example for training the DAL model and present the classification results alongside comparison methods. Considering the space limitation and the simplicity of the demonstration, for the comparison methods, we only show one domain adaptation method TCA, and one traditional supervised learning method LSSVM.
Specifically, the toy example in Figure \ref{Toy example}, as depicted in the first row, encompasses four distinct classification tasks characterized by diverse data distributions. Task 0 represents the initiation stage and involves fully labeled data for training the source model. Tasks 1 to 3 correspond to evolving task flow. The experimental results are displayed in Figure \ref{Toy example} and we have the following observations. 
\begin{itemize}
	\item As seen from the results in Figure \ref{Toy example}, compared with the comparison methods, the DAL method exhibits significant performance advantages. The classification results are closest to the first row, which represents the optimal classifier. The reason may be that DAL has inherited the model trained in the former task in a reasonable way, that is, the optimal transport based on Encoding Feature Marginal Distribution Information (EFMDI) enables transport model reuse across different domains.
	\item For domain adaptation methods, while they achieve acceptable performance in task 2, their effectiveness significantly diminishes as the task flow progresses due to a dearth of annotated source domain data in evolving data distribution scenarios. As shown in the figure, TCA almost fails on Task3 and Task4.
	\item The traditional supervised learning method, LSSVM, is trained with limited current labeled data, which exhibits slightly superior performance compared to the domain adaptation method as the task flow evolves. This advantage stems from the utilization of a small amount of labeled data during the learning of the model classifier and the ease of the binary classification task. However, due to insufficient labeled data, the model's performance remains consistently unsatisfactory. 
\end{itemize}
\renewcommand\arraystretch{1}
\renewcommand\tabcolsep{1pt} %列间距
\begin{table*}[!t]
	\footnotesize
	\caption{Comparisons of classification performance (test accuracy, mean(std)). The best performance and its comparable performances based on paired t-tests at a 95\% significance level are highlighted in boldface.}
	\label{Benchmark}
	\centering
	\begin{tabular}{c|c|ccccccc|cc}
		\toprule[0.75pt]
		\midrule[.5pt]
		Dataset & Task & TCA & JDA  & DTLS  & W-BDA  & LS-OT & LS-G & LSSVM & DAL-LS & DAL-CEL\\
		\midrule[ .5pt]
		\multirow{4}{*}{Electricity}
		& Task1  &.654(.003)&.667(.013)&.691(.013)&.652(.015)&.777(.017)&.545(.011)&.544(.029)&.811(.012)& \textbf{.831(.015)}\\
		& Task2  &.572(.019)&.571(.024)&.669(.018)&.587(.029)&.540(.008)&.516(.012)&.515(.046)& \textbf{.782(.008)}&.719(.018)\\
		& Task3  &.569(.021)&.582(.019)&.557(.016)&.558(.018)&.742(.017)&.586(.015)&.703(.035)&.787(.009)& \textbf{.791(.011)}\\
		& Task4  &.552(.008)&.526(.017)&.497(.018)&.550(.016)&.710(.021)&.600(.019)&.600(.036)&.767(.012)& \textbf{.770(.012)}\\
		\midrule[.5pt]
		
		\multirow{4}{*}{Weather}
		& Task1  &.753(.033)&.768(.044)&.765(.017)&.755(.021)&.719(.038)&.719(.018)&.665(.065)& \textbf{.788(.022)}&.717(.020)\\
		& Task2  &.713(.016)&.716(.015)&.717(.036)&.712(.025)&.688(.019)&.688(.018)&.685(.050)& \textbf{.722(.020)}&.689(.027)\\
		& Task3  &.733(.022)&.741(.021)&.719(.011)&.742(.022)&.695(.011)&.698(.022)&.739(.048)& \textbf{.790(.019)}&.698(.036)\\
		& Task4  &.682(.024)&.715(.025)&.707(.010)&.710(.019)&.688(.040)&.689(.011)&.647(.037)& \textbf{.756(.020)}&.694(.025)\\
		\midrule[.5pt]
		
		\multirow{4}{*}{Spam}
		& Task1  & \textbf{.934(.016)}&.914(.020)&.932(.015)&.913(.010)&.765(.039)&.765(.032)&.870(.038)&.765(.036)&.763(.035)\\
		& Task2  &.878(.010)&.837(.014)& \textbf{.928(.018)}&.885(.042)&.822(.018)&.822(.025)&.853(.038)&.822(.025)&.817(.021)\\
		& Task3  & \textbf{.909(.016)}&.877(.028)&.824(.012)&.886(.011)&.765(.011)&.764(.019)&.827(.044)&.764(.009)&.762(.021)\\
		& Task4  & \textbf{.914(.015)}&.894(.008)&.862(.022)&.886(.023)&.773(.007)&.787(.005)&.771(.016)&.763(.033)&.756(.008)\\
		\midrule[.5pt]
		
		\multirow{4}{*}{PIE}
		& Task1  &.112(.007)&.129(.015)&.357(.041)&.149(.023)&.610(.010)&.028(.005)&.247(.038)&.652(.018)& \textbf{.666(.012)}\\
		& Task2  &.056(.005)&.070(.016)&.172(.012)&.055(.007)&.628(.008)&.021(.007)&.243(.043)&.646(.018)& \textbf{.709(.012)}\\
		& Task3  &.047(.003)&.060(.012)&.100(.003)&.044(.009)&.619(.016)&.024(.004)&.437(.038)& \textbf{.829(.011)}&.772(.006)\\
		& Task4  &.023(.007)&.041(.017)&.069(.010)&.040(.015)&.543(.019)&.028(.007)&.233(.046)&.660(.016)& \textbf{.673(.007)}\\
		\midrule[.5pt]
		
		\multirow{3}{*}{VLSC}
		& Task1  &.543(.015)&.541(.012)&.483(.037)&.483(.038)&.600(.005)&.331(.025)&.682(.032)& \textbf{.703(.014)}&.399(.006)\\
		& Task2  &.402(.024)&.401(.029)&.390(.028)&.390(.028)&.616(.019)&.464(.015)&.609(.048)& \textbf{.633(.013)}&.441(.014)\\
		& Task3  &.293(.095)&.332(.027)&.304(.026)&.304(.053)&.845(.021)&.617(.022)&.769(.042)& \textbf{.861(.005)}&.469(.011)\\
		\midrule[.5pt]
		
		\multirow{3}{*}{AR}
		& Task1  &.616(.022)&.609(.020)&.743(.014)&.611(.014)&.601(.014)&.502(.027)&.678(.045)&.770(.014)& \textbf{.817(.016)}\\
		& Task2  &.555(.010)&.557(.032)&.597(.030)&.553(.019)&.612(.034)&.500(.042)&.499(.057)&.711(.022)& \textbf{.744(.033)}\\
		& Task3  &.527(.026)&.553(.021)&.555(.037)&.555(.027)&.529(.025)&.499(.030)&.557(.051)&.711(.019)& \textbf{.723(.020)}\\
		\midrule[.5pt]
		
		\multirow{3}{*}{AWDC}
		& Task1  &.325(.057)&.374(.020)&.331(.037)&.330(.049)&.320(.035)&.115(.020)&.131(.049)& \textbf{.487(.030)}&.486(.030)\\
		& Task2  &.244(.033)&.237(.017)&.115(.035)&.254(.072)&.342(.030)&.107(.033)&.167(.105)&.436(.084)& \textbf{.540(.016)}\\
		& Task3  &.178(.035)&.223(.064)&.102(.058)&.197(.041)&.100(.008)&.130(.064)&.180(.161)& \textbf{.388(.073)}&.100(.014)\\
		\midrule[.5pt]
		
		\multirow{2}{*}{C-DT}
		& Task1  &.426(.024)&.450(.018)&.430(.019)&.243(.028)&.248(.010)&.029(.007)&.376(.032)& \textbf{.597(.021)}&.025(.003)\\
		& Task2  &.122(.012)&.123(.016)&.102(.013)&.049(.012)&.135(.012)&.025(.007)&.150(.029)& \textbf{.229(.015)}&.025(.002)\\
		\midrule[.5pt]
		
		\multirow{4}{*}{Magic}
		& Task1  &.696(.027)&.712(.039)&.726(.020)&.686(.034)&.640(.031)&.643(.011)&.642(.048)& \textbf{.791(.016)}&.742(.017)\\
		& Task2  &.679(.030)&.693(.021)&.709(.019)&.680(.011)&.651(.025)&.647(.031)&.648(.059)& \textbf{.778(.023)}&.729(.018)\\
		& Task3  &.678(.024)&.681(.018)&.702(.022)&.653(.021)&.642(.025)&.639(.019)&.644(.036)& \textbf{.761(.017)}&.714(.042)\\
		& Task4  &.678(.025)&.683(.023)&.699(.019)&.658(.030)&.637(.025)&.643(.020)&.640(.048)& \textbf{.795(.014)}&.723(.032)\\
		\midrule[.5pt]
		
		\multirow{4}{*}{Poker hand}
		& Task1  &.444(.023)&.445(.007)&.436(.016)&.435(.016)&.454(.011)&.516(.012)&.404(.035)& \textbf{.519(.018)}&.506(.006)\\
		& Task2  &.443(.010)&.445(.017)&.437(.011)&.450(.015)&.504(.018)&.501(.022)&.467(.031)& \textbf{.517(.010)}&.407(.010)\\
		& Task3  &.440(.019)&.450(.008)&.445(.014)&.450(.007)& \textbf{.509(.015)}&.502(.009)&.500(.036)& \textbf{.504(.014)}&.499(.010)\\
		& Task4  &.447(.025)&.450(.012)&.441(.018)&.443(.020)& \textbf{.510(.006)}&.500(.013)&.450(.032)&.507(.024)& \textbf{.510(.017)}\\
		\midrule[.5pt]
		
		\multirow{4}{*}{Waveform}
		& Task1  &.688(.042)&.734(.031)&.706(.030)&.707(.018)&.754(.030)&.611(.048)&.698(.049)&.763(.028)& \textbf{.800(.041)}\\
		& Task2  &.686(.033)&.712(.028)&.706(.039)&.708(.017)&.734(.020)&.613(.020)&.701(.051)&.735(.048)& \textbf{.790(.028)}\\
		& Task3  &.660(.025)&.698(.050)&.674(.020)&.717(.053)&.754(.032)&.595(.027)&.681(.056)&.716(.020)& \textbf{.834(.021)}\\
		& Task4  &.659(.058)&.687(.016)&.705(.016)&.683(.028)&.665(.042)&.572(.046)&.574(.071)&.775(.030)& \textbf{.792(.026)}\\
		\midrule[.5pt]
		\bottomrule[.75pt]
	\end{tabular}
\end{table*}

\subsection{Benchmark Datasets}\label{s6.3}

In this case, similar to the setup above, we evaluate the performance of our DAL approach with other closely related methods on several benchmark datasets. We report the Acc results in Table \ref{Benchmark}. There are several observations as follows.
\begin{itemize}
	\item The performance of our DAL-LS and DAL-CEL is still better than other methods in most cases. For example, as shown by the $t$-test results in boldface, the numbers of win/tie/loss of our methods are 33, 2, and 4 respectively. 
	\item Compared to the results of domain adaptation methods, we can see that the performance of the domain adaptation method shows a decreasing trend with the progress of the task flow, especially when the classification task has many categories, such as the \textbf{PIE} dataset with 68 categories. The possible reason is that using the prediction of the former task as the source domain label of the current task may result in error accumulation, consequently leading to a continuous deterioration of performance. In contrast, the performance of the DAL method shows some fluctuation, which may be due to the different classification difficulty of different tasks. But the best performance is still achieved in most cases.
	\item Comparing the DAL-LS, LS-0T, and LS-G results, we can draw two conclusions. First of all, compared with the complete model, removing any one of the two terms transport model reuse or manifold regularization will lead to performance degeneration, which shows that each term does contribute to the performance improvement of DAL. Secondly, among these two terms, the transport model reuse plays a more important role. This proves that the main purpose of the introduction of transport model reuse is to rectify a closely related model by optimal transport and then enable model reuse across diverse data distributions, which is consistent with the original design intention of the DAL model.
\end{itemize}  

\subsection{Performance Analyses} \label{s6.4}

We further analyze the performance of our proposed DAL method from different aspects in this subsection. Firstly, we explore the influence of the evolving data distribution. Secondly, the value of the objective function is shown to validate the results in Proposition \ref{Proposition3}.

\subsubsection{Influence of evolving data distribution}

\begin{figure}[t]
	\centering
	\subfigure[Electricity]{
		\includegraphics[width=13cm]{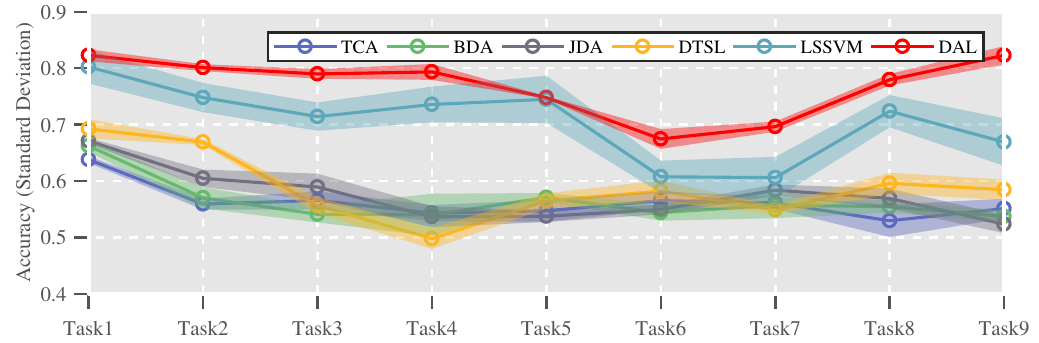}}
	\subfigure[Weather]{
		\includegraphics[width=13cm]{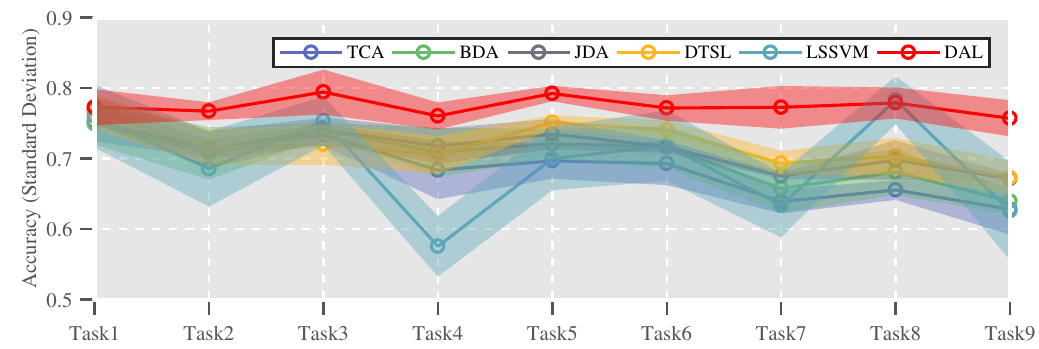}}
	\caption{The influence of evolving data distribution on the performance of each model.}
	\label{Influence}
	\vskip -.1in
\end{figure}

Since classification tasks arrive as streams with evolving data distributions, we would like to show the influence of evolving data distribution on the performance of each method. We conduct experiments on two data sets, i.e., Electricity, Weather. The data of these two data sets are sorted according to their arrival time and their distribution gradually changes as time passes, so these two dataset represents gradual distribution evolution. Specifically, each dataset is partitioned into 10 tasks based on the arrival time of data, while keeping other experimental settings consistent with the previous main experiment. As shown in Figure. \ref{Influence}, the experimental results on these two datasets are reported. We have the following observations.
\begin{itemize}
	\item Our proposed DAL method still performs better than other methods in most cases throughout the data distribution evolution. As the task flow progresses, the advantages of DAL become increasingly evident, thereby demonstrating the effectiveness of DAL in tracking the evolving data distribution.
	\item For the LSSVM method, the performance of the LSSVM shows large volatility throughout the evolution process. In contrast, DAL exhibits a higher level of stability in performing the two task flows, thereby confirming the effectiveness of incorporating transport model reuse and manifold regularization to enhance both the performance and stability of the model.
	\item For the domain adaptation methods, their performances show a decreasing trend as the task flow progresses under both two evolving data distributions. The reason may be that using the prediction result of the former task as the source domain labels of the current task will cause the accumulation of errors and further lead to the continuous degradation of performance. 
\end{itemize}
\begin{figure}[H]
	\begin{center}
		\subfigure[Electricity]{
			\includegraphics[width=.4\columnwidth]{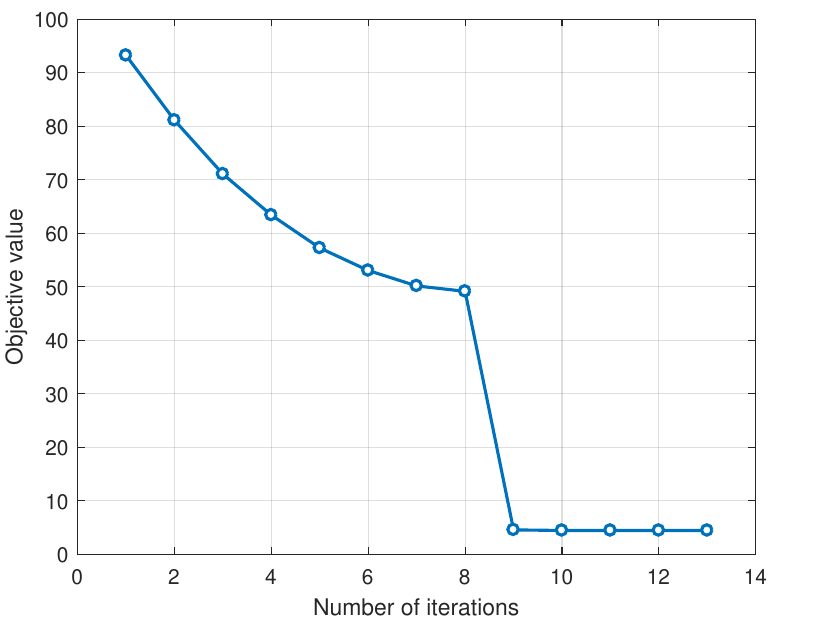}}
		\label{}
		\subfigure[Waveform]{
			\includegraphics[width=.4\columnwidth]{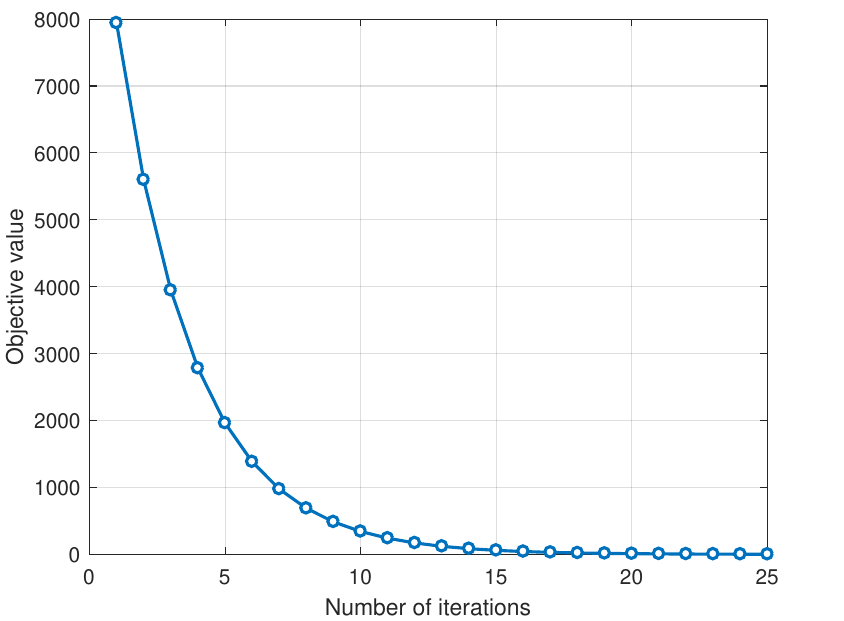}}
		\label{}
		\subfigure[Poker hand]{
			\includegraphics[width=.4\columnwidth]{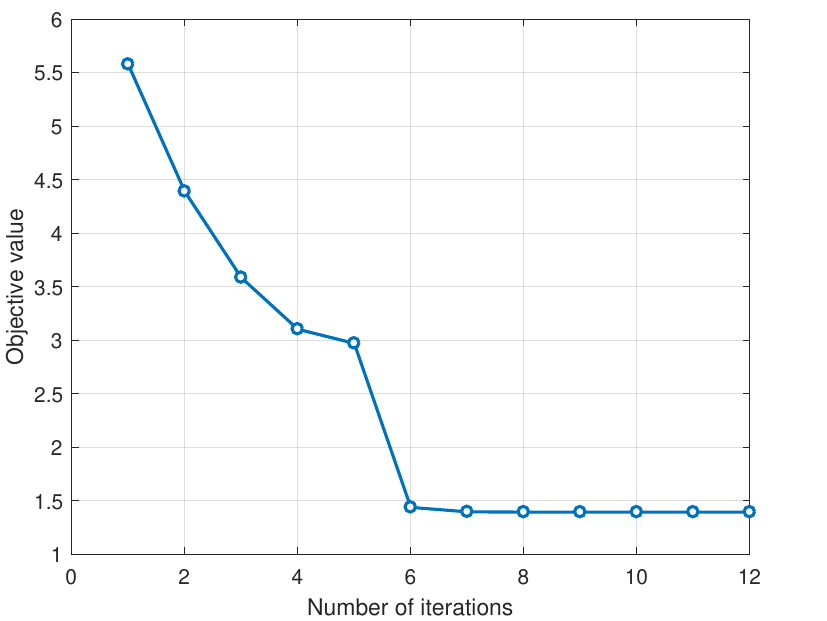}}
		\label{}
		\subfigure[Weather]{
			\includegraphics[width=.4\columnwidth]{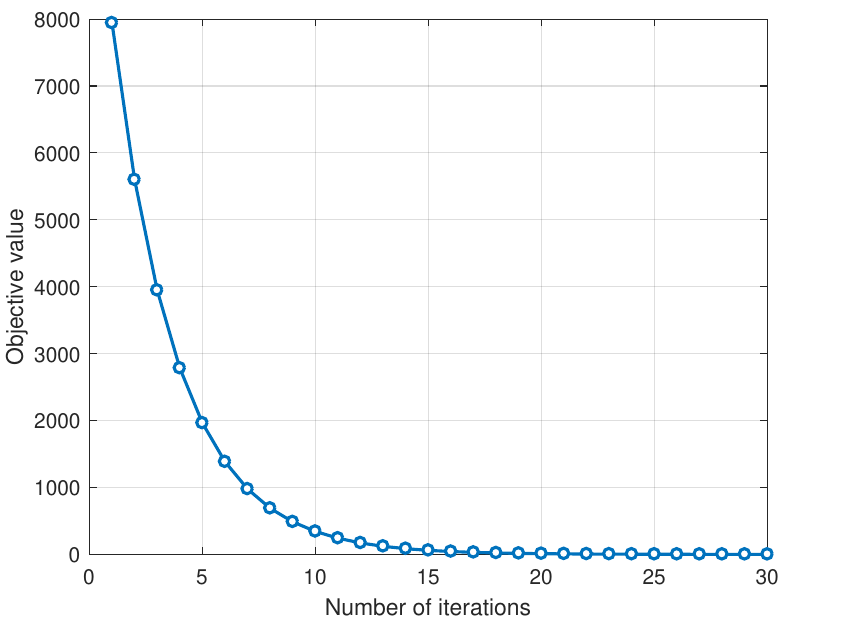}}
		\label{}
	\end{center}
	\caption{The convergence behavior of DAL-CEL on four datasets.}
	\label{Convergence}
	\vskip -.1in
\end{figure}

\subsubsection{Convergence Behavior}\label{s6.4}

It can be seen from Algorithm \ref{alg1}, our proposed model DAL-CEL is solved in an alternative way. The result in Proposition \ref{Proposition3} has shown its convergence theoretically. For demonstration, we also show the objective function values of a local step in the evolution process on four datasets, i.e., \textbf{Electricity, Waveform, Poker hand} and \textbf{Weather}. The figures are presented in Fig. \ref{Convergence}, Besides, we also use the logarithmic scale in the $y$-axis to show the sensitive changes. Obviously, our optimization algorithm converges in a fast way and the iteration number is often less than 20.

\subsection{Application to Digit Recognition}\label{s6.5}

\begin{figure}[h]
	\setlength{\abovecaptionskip}{-.1cm}
	\setlength{\belowcaptionskip}{-.1cm}
	\begin{center}
		\includegraphics[width=13cm]{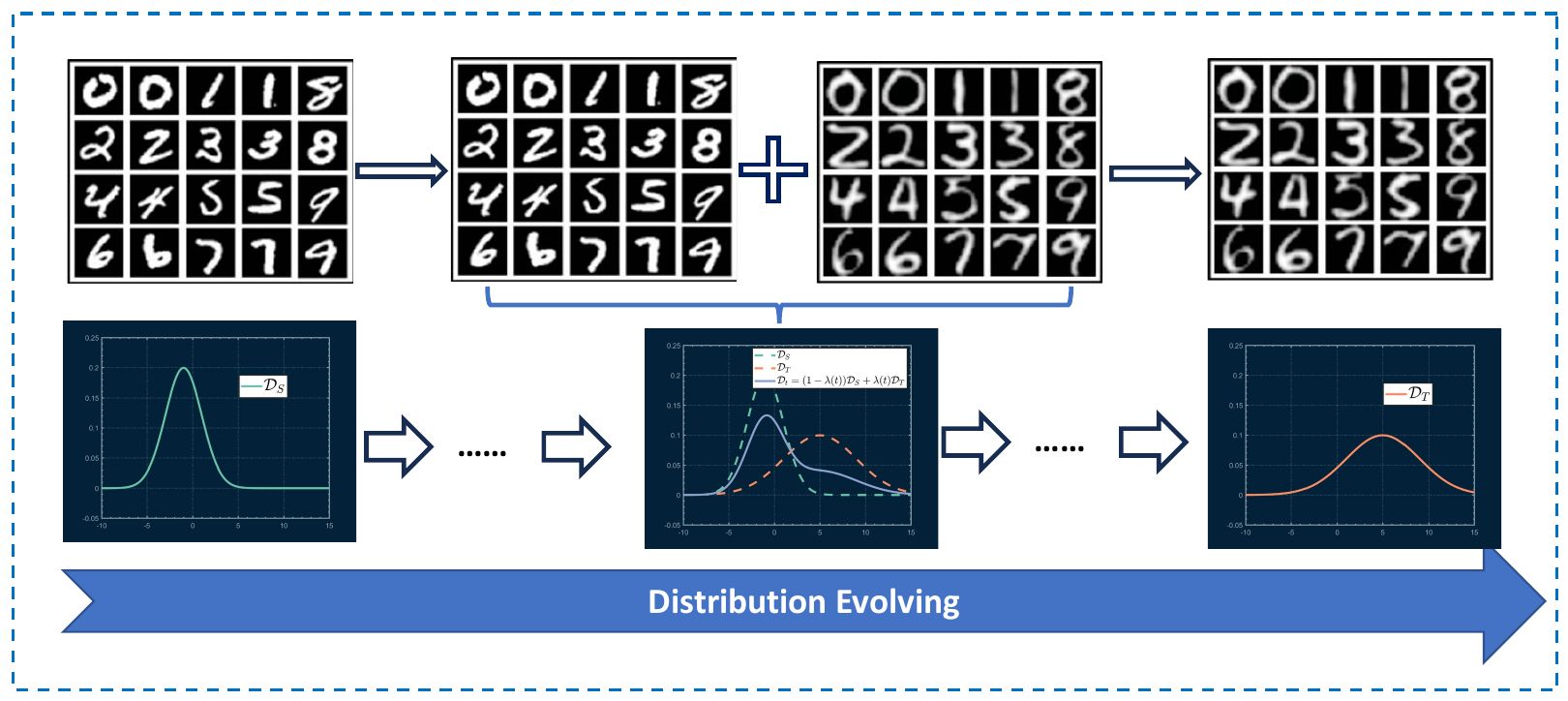}
	\end{center}
	\caption{Illustration of the evolving data distribution classification tasks from \textbf{Mnist} to \textbf{USPS}. In the figure, the distribution of \textbf{Mnist} is denoted as $\mathcal{D}_S$ (green curve) on the left, the distribution of \textbf{USPS} is denoted as $\mathcal{D}_T$ (orange curve) on the right, the mixture distribution of $\mathcal{D}_S$ and $\mathcal{D}_T$ is denoted as $\mathcal{D}_{t}$ (blue curve) in the middle. $\lambda(t)$ is the time-dependent mixture proportion, here we take $\lambda(t)=0.4$ as a demonstration.}
	\label{Digit Recognition}
	\vskip -.1in
\end{figure}

In this subsection, we show the effectiveness of our proposed methods on a real-world Digit Recognition task. Consider two typical handwritten digit pictures datasets \textbf{Mnist} and \textbf{USPS}, \textbf{USPS} dataset consists of 9298 images and \textbf{Mnist} dataset consists of 70,000 images. As shown in Figure \ref{Digit Recognition}, they share 10 classes of digits and follow very different distributions. 

Looking back at the scenario in this article, we would like to show the performance of each model under an evolving data distribution. Inspired by the shift assumption in investigation \cite{DBLP:conf/nips/KiryoNPS17,DBLP:conf/nips/Zhang0MZ20}, we formulate the evolving data distribution as follows. 
\begin{equation}\label{evolving distribution}
\mathcal{D}_{t} = (1-\lambda(t))\mathcal{D}_S+\lambda(t)\mathcal{D}_T,
\end{equation}
where $\lambda(t)$ is a time-dependent mixture proportion, $\lambda(0)=0$ and $\lambda(1)=1$. $\mathcal{D}_S$ and $\mathcal{D}_T$ are two different data distributions. In our experiment, we select all samples of \textbf{USPS} and a subset Mnist with 10,000 samples. Specifically, denote the distribution of \textbf{Mnist} as $\mathcal{D}_S$ and the distribution of \textbf{USPS} as $\mathcal{D}_T$, and take the value of $\lambda(t)$ as $0, .2, .4,\cdots,1$, respectively. And then we obtain a task flow $\textbf{Mnist} \to \textbf{USPS}$ with 6 tasks with evolving data distribution. Switch source/target
pair to get another task flow $\textbf{USPS} \to \textbf{Mnist}$.

\begin{figure*}[h]
	\centering
	\subfigure[$\textbf{Mnist} \to \textbf{USPS}$]{
		\includegraphics[width=13.5cm]{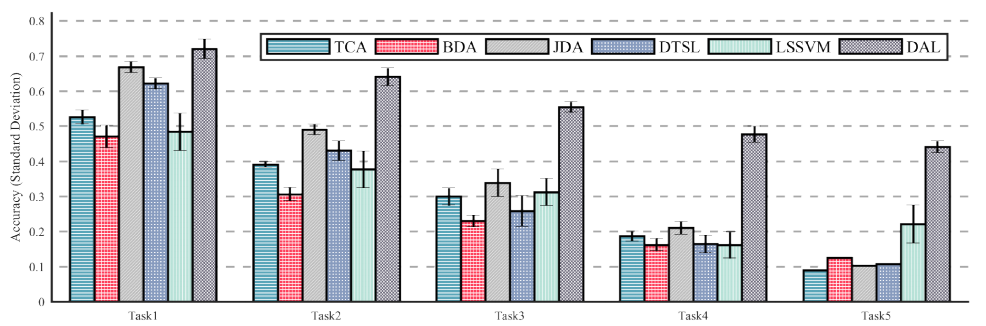}}
	\caption{Comparisons of classification performance (test accuracy, mean(std)) of DAL and representative comparison methods on the evolving digit recognition task of $\textbf{Mnist} \to \textbf{USPS}$.}
	\label{Task Flow}
\end{figure*}

Similarly, we compare DAL with domain adaption methods and LSSVM and keep the experimental setup the same as in Section \ref{s6.1}. As shown in Table. \ref{UVM} and Figure. \ref{Task Flow}, the experimental results of two evolving data distributions $\textbf{Mnist} \to \textbf{USPS}$ and $\textbf{USPS} \to \textbf{Mnist}$ are reported. We have the following observations.
\begin{itemize}
	\item Our proposed DAL method always outperforms other baseline methods in this real application throughout the data distribution evolution. It verifies the application potential of our proposals.
	\item The DAL method demonstrates superior stability in comparison to the other methods under consideration, owing to the incorporation of two crucial regularizations: \textbf{Transport Model Reuse} and \textbf{Manifold Regularization}.
	\item In the digit recognition task, the performance of the domain adaptation method shows a downward trend with the advancement of the task flow, which is consistent with the conclusion of the previous experiment. 
\end{itemize}
\renewcommand\arraystretch{0.9}
\renewcommand\tabcolsep{1pt} %列间距
\begin{table*}[h]
	\caption{Acc(mean(std)) comparison on the evolving digit recognition task of $\textbf{USPS} \to \textbf{Mnist}$. The best performance and its comparable performances based on paired t-tests at a 95\% significance level are highlighted in boldface.}
	\label{UVM}
	\centering
	\begin{tabular}{c|c|ccccc|c}
		\toprule[0.75pt]
		\midrule[.5pt]
		Evolution & Tasks & TCA & JDA  & W-BDA  & DTLS  &LSSVM & DAL\\
		\midrule[.5pt]
		\multirow{5}{*}{$\textbf{USPS} \to \textbf{Mnist}$}
		& Task1  & .277(.018)& .364(.013)& .255(.010)& .337(.015)& .220(.047)& \textbf{.590(.027)}\\
		& Task2  & .232(.017)& .273(.018)& .217(.011)& .239(.030)& .210(.048)& \textbf{.561(.015)}\\
		& Task3  & .192(.019)& .223(.011)& .157(.017)& .187(.032)& .373(.050)& \textbf{.567(.030)}\\
		& Task4  & .138(.013)& .223(.009)& .136(.015)& .184(.020)& .269(.051)& \textbf{.601(.013)}\\
		& Task5  & .114(.000)& .214(.000)& .120(.000)& .184(.000)& .582(.047)& \textbf{.767(.025)}\\
		\midrule[.5pt]
		\bottomrule[.75pt]
	\end{tabular}
\end{table*}
	
\section{Conclusion}\label{s7}

We have studied the problem of classification under evolving data distribution, which is rarely studied and of great importance. To enable the model to track the evolving data distribution, we formulate a new framework named Distribution Adaptable Learning (DAL). The DAL framework is particularly useful for robust learning in compound cases in an open and dynamic environment. DAL has a solid theoretical guarantee, together with a smart optimization strategy. In practice, our approach is more beneficial to handle the evolution of data distribution since we achieved model reuse across different distributions through transport model reuse. To eliminate the obstruction to model reuse caused by distribution evolution, we proposed a new strategy named Encode Marginal Distribution InformaTion of features (EMDI), which expands the scope of optimal transport to enable model reuse across diverse data distributions, thereby enhancing the reusable and evolvable properties of DAL in accommodating evolving data distributions. In this paper, we only designed one strategy by adding regularizers for model inheriting and reusing. How to incorporate other strategies and approaches is an interesting future study. Besides, how to efficiently learn the optimal transport matrix across different feature domains is also worth studying. 

\section*{Acknowledgments}

This work was partially supported by the National Key Research and Development Program (No. 2022ZD0114803), the Key NSF of China under Grant No. 62136005, and the NSF of China under Grant No. 61922087. Chenping Hou and Yuhua Qian are the corresponding authors.

\bibliography{mybib}

\end{document}